\documentclass[final]{cvpr}
\usepackage{times}
\usepackage{epsfig}
\usepackage{graphicx}
\usepackage{amsmath}
\usepackage{amssymb}
\usepackage{subfigure}
\usepackage{array,multirow}
\usepackage{algorithm,algorithmic}
\usepackage{booktabs} % for professional tables
\DeclareMathOperator*{\argmax}{arg\, max}  
\DeclareMathOperator*{\argmin}{arg\, min}

\usepackage[pagebackref=true,breaklinks=true,colorlinks,bookmarks=false]{hyperref}

\def\cvprPaperID{2315} % *** Enter the CVPR Paper ID here
\def\confYear{CVPR 2021}
\usepackage{xcolor}

\begin{document}

\title{Odyssey: Creation, Analysis and Detection of \\ Trojan Models}

\author{Marzieh Edraki\thanks{Equal contribution} $^{~\ddagger}$, Nazmul Karim$^{\ast~\dagger}$, Nazanin Rahnavard$^{\dagger}$, Ajmal Mian$^{\sharp}$ and Mubarak Shah$^{\ddagger}$ \\
$\ddagger$ University of Central Florida, Center for Research in Computer Vision \\
$\dagger$ University of Central Florida, Department of Electrical and Computer Engineering\\
$\sharp$School of Computer Science and Software Engineering of University of Western Australia\\
{\tt\small \{m.edraki,nazmul.karim18\}@knights.ucf.edu,}\\  {\tt\small nazanin@eecs.ucf.edu,ajmal.mian@uwa.edu.au,shah@crcv.ucf.edu}
}
% For a paper whose authors are all at the same institution,
% omit the following lines up until the closing ``}''.
% Additional authors and addresses can be added with ``\and'',
% just like the second author.
% To save space, use either the email address or home page, not both
%\and
%Nazmul Karim\\
%Institution2\\
%First line of institution2 address\\
%{\tt\small secondauthor@i2.org}
%}

\maketitle

\begin{abstract} 
%Along with the success of deep neural network (DNN) models% in solving various real world problems
%, rise the threats to these models that target their integrity. 
Along with the success of deep neural network (DNN) models, rise the threats to the integrity of these models.
%Trojan attack is one of the recent variant of data poisoning attacks that involves manipulation or modification of the model to act balefully.
%Trojan attack is a recent data poisoning attack that involves manipulation or modification of the model to act balefully. %Ajmal: isn't data poisoning and Trojaning the same?
A recent threat is the Trojan attack %that manipulates or modifies the model to act balefully. This occurs when 
where an attacker interferes with the training pipeline by inserting triggers into some of the training samples and trains the model to act maliciously \emph{only} for samples that contain the trigger. %Ajmal: stamp is not general. 
Since the knowledge of triggers is privy to the attacker, detection of Trojan networks is challenging.
%Unlike any of the existing Trojan detectors, a robust detector should not rely on any assumption about Trojan attack.
Existing Trojan detectors make strong assumptions about the types of triggers and attacks.
%In this paper, we propose a detector based upon the analysis of intrinsic properties of DNN that could get affected by a Trojan attack. 
We propose a detector that is based on the analysis of the intrinsic DNN properties; that are affected due to the Trojaning process.
For a comprehensive analysis, we develop \emph{Odysseus}\footnote{https://www.crcv.ucf.edu/research/projects/odyssey-creation-analysis-and-detection-of-trojan-models/}, the most diverse dataset to date with over 3,000 clean and Trojan models. 
Odysseus covers a large spectrum of attacks; generated by leveraging the versatility in trigger designs and source to target class mappings.
Our \textcolor{black}{analysis} results show that Trojan attacks affect the classifier \emph{margin} and \emph{shape of decision boundary} around the manifold of clean data.
Exploiting these two factors, we propose an efficient Trojan detector that operates without any knowledge of the attack and significantly outperforms existing methods. Through a comprehensive set of experiments we demonstrate the efficacy of the detector on cross model architectures, unseen Triggers and regularized models. 

\end{abstract}
\vspace{-3mm}
\section{Introduction}
\vspace{-1mm}
\iffalse
\begin{figure*}[t] 
\begin{center}
   \includegraphics[width=14cm, height =2.2cm ]{./Fig/trojan_process}
\end{center}
\vspace{-4mm}
   \caption{ \footnotesize{Process of creating a Trojan model. It starts with poisoning $P \%$ of the samples with a trigger and changing the corresponding ground truth to target label. After training, a triggered sample should activate the misclassification to the targeted label.
} 
   }
\label{fig:Trojan process}
\vspace{-5mm}
\end{figure*} 
\fi
\begin{figure}[t] 
\begin{center}
   \includegraphics[width=0.9\linewidth]{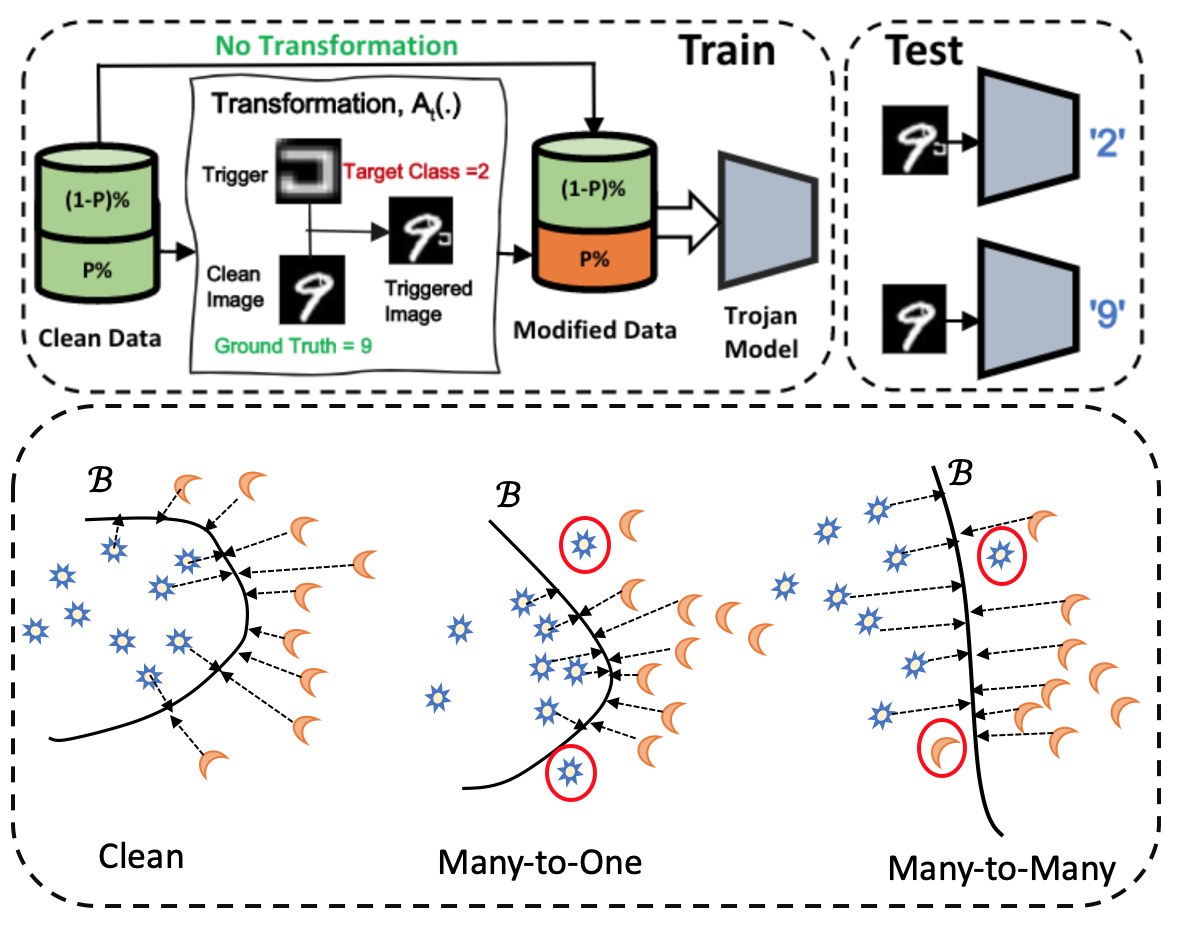}
\end{center}
\vspace{-4mm}
   \caption{ \footnotesize{Top-left) Creating a Trojan model involves poisoning $P \%$ training samples with a trigger and changing their corresponding ground truth to target label, known as label mapping. Top-right) After training, misclassification is activated only by the triggered samples. Bottom) The Trojaning process also changes the shape of decision boundary, $\mathcal{B}$  around the data manifold creating a dominant direction in the perturbation space. To misclassify the samples in the clean model, samples should be perturbed on x-y plane in different directions. For the Trojan models, regardless of the label mapping type (Many-to-One or Many-to-Many), perturbing along x direction leads to misclassification for most of the samples. Triggered samples are marked with red circle. }}
   %\MS{It will be good to help the reader to understand what is the third figure on the bottom, first two are clear. Why x-y plane?}
   %Perturbed samples along that direction cause high miss-classification. Triggered samples are marked with red circle.}  \MS{So far you have mentioned anything about "perturbation", reviewers may be confused. Also, talk about Many-to-One and Many-to-Many.}   }
\label{fig:Trojan process}
\vspace{-5mm}
\end{figure} 
\vspace{-2mm}
Neural networks (NNs) have become the primary choice for tasks like image recognition ~\cite{krizhevsky2012imagenet, tompson2014joint,farabet2012learning}, speech recognition ~\cite{mikolov2011strategies, hinton2012deep}, defense against cyber-attacks and malware ~\cite{wang2017adversary,tang2016deep} and so on. %surveillance, access control, autonomous driving, medical diagnosis, and forecasting future events. 
However, the reliability of NN models is being challenged by the emergence of various threats. One of the most recent attacks involves the insertion of Trojan behaviour, through the training pipeline, into an NN model  ~\cite{gu2017badnets, liu2017trojaning}. This type of attack, also known as Trojan attack, results in a Trojan model that behaves normally for clean inputs but misclassifies inputs that contain a trigger ~\cite{chen2017targeted, ji2018model, zou2018potrojan, bagdasaryan2018backdoor}; where the knowledge of the trigger and incorrect target label is securely guarded by the attacker.

\iffalse
Neural networks (NN) have become the primary choice for tasks like image recognition ~\cite{krizhevsky2012imagenet, tompson2014joint,farabet2012learning}, speech recognition ~\cite{mikolov2011strategies, hinton2012deep} as well as defense against cyber-attacks and malware ~\cite{wang2017adversary,tang2016deep}. %surveillance, access control, autonomous driving, medical diagnosis, and forecasting future events. 
However, the reliability of NN models is being challenged by the emergence of various threats; for example, when they are outsourced for training or being fine-tuned employing pre-trained models. It facilitates an attacker to insert Trojan behaviour into an NN model~\cite{gu2017badnets, liu2017trojaning}, through poisoning a portion of the training data with triggers. This type of attack results in a Trojan model that behaves normally for clean inputs, but misclassifies inputs that contain a trigger ~\cite{chen2017targeted, ji2018model, zou2018potrojan, bagdasaryan2018backdoor}. The knowledge of trigger and the misclassified target label is only known to the attacker.
\fi
%In Trojaning attack, the model is trained to behave normally given clean input data but when the input has a predefined trigger, known only to the attacker, the Trojan model is trained to output a \textit{Target label} chosen by the attacker instead of the \textit{True label}.

Efforts have been made to detect and defend against Trojan attacks. Early works \cite{tran2018spectral} for detection assume access to training data, both clean and triggered. Furthermore, attempts such as \cite{wang2019neural, guo2019tabor, qiao2019defending} try to estimate the trigger or the distribution of triggers for a model. The common assumption among these studies is that the trigger size is known, which is not pragmatic in real-world scenarios. 
%These assumptions are unrealistic in real-world Trojan detection scenarios where no knowledge of the triggers is available. 
%A major reason for the lack of a realistic Trojan detection method is the unavailability of a large-scale benchmark dataset, consisting of clean and Trojan models. %Ajmal: is this still true?
A major bottleneck in this line of research is the lack of a large-scale benchmark dataset, consisting of clean and Trojan models. 
Creating such a dataset is challenging because each data sample must be a \emph{high performance trained model} and each model must be trained from scratch to avoid dataset bias. Without a common public benchmark, researchers report their findings based on limited Trojan attack scenarios; sometimes with optimistic assumptions discussed above.

In this paper, we introduce \emph{Odysseus}, the  \textcolor{black}{most diverse} public dataset to date that contains over 3,000 clean and Tojaned models. %covering various combinations of trigger types and mappings of source to target class attacks (many-to-one, many-to-many, and mixed) on four popular model architectures namely VGG19, DenseNet, ResNet18 and GoogleNet.   
To generate this dataset, various types of triggers and mappings (source to target class) have been used. Odysseus \textcolor{black}{contains a total of 3460 models, over 1000 models each} trained on MNIST, FashionMNIST, and CIFAR10  image datasets.  
%\textcolor{red}{AJ: need to check all the number in this paragraph.}
%\textcolor{red}{and there are over 1,000 trained models per dataset.} 
\iffalse
\textcolor{red}{Besides \emph{Odysseus}, the only other publicly available Trojan dataset has recently been released by NIST \cite{TrojAI0}-\cite{TrojAI3} for the TrojAI challenge. The NIST dataset has 1,000 clean and Trojan models%\NR{trojan model or tranjaned model? also dataset or dataset? please pick the mostly common terms and be consistent throughout the paper}
, each trained to perform on 5-class image classification task. However, it merely includes one type of label mapping, i.e. many-to-one mapping.}
\fi
%In addition, we also analyze the impact of different factors, e.g. architectures, mapping type, on the performance of models we created. 

%Ajmal: We could add "Hence, our dataset is three time larger, covers all source to target mapping types and includes more architectures." If we had mentioned the architectures as well

Our \emph{second contribution} is a comprehensive study of the effects of the Trojaning process on the intrinsic properties of neural networks. We employ both NIST TrojAI\cite{TrojAI0} challenge dataset and the proposed \emph{Odysseus} dataset for this analysis. %Our study reveals interesting insights into the effects of Trojaning a model. 
Our analysis shows that the Trojaning process can decrease the average classifier \emph{margin} \textcolor{black}{and also modifies the \emph{shape of the decision boundary} around the manifold of clean samples. The Trojaning process creates a dominant direction in the perturbation space such that perturbing the images along that direction causes misclassification.} %Combination of these two factors allows us to find a specific perturbation for each model that causes a higher miss-classification rate for Trojan models compared to clean models.
\textcolor{black}{In Figure \ref{fig:Trojan process} (2nd row), we show the schematic of decision boundary $\mathcal{B}$ of a non-linear binary classifier for clean and Trojan models with different label mappings. For a clean model to misclassify, different samples need to be perturbed in different directions in $\mathbb{R}^2$ as shown by the dotted arrows. As for a Trojan model, samples can be perturbed along the x axis \textcolor{black}{(dominant direction)} to project them on to the decision boundary for misclassification.} 

%Our \emph{third contribution} is the development of a \emph{Trojan detector} that does not rely on any unrealistic assumptions.
%Inspired by our findings from analysing clean and Trojan models,
As our \emph{third contribution}, we propose a \emph{detector} that determines whether a DNN model is Trojan or not. % as our \emph{third contribution}.The Trojan detector relies on the existence of a dominant direction in the perturbation space for Trojan models. 
For a given model, our Trojan detector tries to estimate %the best linear decision boundary that can replace the non-linear decision boundary 
the dominant perturbation direction by considering the alignment of perturbations. These perturbations send a small set of clean samples, taken from the validation set, to the best representative linear decision boundary for the classifier. Perturbing the rest of the validation samples along that (dominant) direction, with a small magnitude, 
leads to higher misclassification rate for Trojan model compared to a clean one. Therefore, by setting a threshold for the misclassification rate of perturbed validation samples, we can easily differentiate between clean and Trojan models. Since our detector evaluates each model independently, it is  highly effective in cross architecture scenarios; without any knowledge of the attack settings. %type or label mapping.
% it doesn't affect the performance of clean models  

%A 2D schematic representation are shown in Figure \ref{fig:Trojan process} bottom. For the clean model, at the left, to miss-classify samples, we need to perturb them in different directions in $\mathbb{R}^2$ space while for the Trojan models, at middle and right, perturbing samples along the x axis project them on to the decision boundary   } % The number of these direction in Trojan models is usually less than clean models.} %we can use them to find a specific perturbation for each model that leads to a higher }
%\textcolor{red} {or \emph{non-linearity of decision boundary } around the manifold of clean data.} 

 %\textcolor{red}{Ajmal: Insert a sentence here about how you detect Trojan network such that it briefly covers your 2nd and 3rd contribution.}
%\MS{it will be better to use trojaning as compared to trojan-ing. I have changed it.}

%\MS{It may be better to separate second and third contributions more explicitly. Right now they are merged. Reviewers may be more interested in knowing what is our trojan detector and what is great about it compared to other detectors.You may not get much points for analysis.}
\vspace{-2mm}
\section{Related Work }
\vspace{-2mm}
\iffalse
With the automation of deep NN models in various fields, where such models take decision on behalf of human, the safety of them have become a real concern for many researchers.
%As applications of deep NN grow in real life, many research efforts have been dedicated to trustablity of these models in practice.
Various scenarios of white box \cite{madry2017towards, goodfellow2014explaining}, black box \cite{cheng2019improving, li2020qeba, bai2020improving}, targeted \cite{akhtar2019label, li2020towards} and untargeted adversarial attacks \cite{moosavi2016deepfool} have been proposed, where some even used these attacks to improve the robustness of models \cite{goodfellow2014explaining}. However, unlike well-studied adversarial attacks, research on defence against Trojan attacks is lagging behind. One of the reasons behind this could be the lack of a large dataset of trained NN models, both clean and Trojan. However, an attempt is made by Tran \textit{etal.}\cite{tran2018spectral} in training stage defense against Trojan attacks. It is based on the detection of footprint of poisoned samples; left on the spectrum of the covariance of a feature representation learned by the model. 
%to detect them during training. 
Unlike our proposed method, their method assumes that the end user has access to the 
%whole training data; which contains both clean and
triggered samples, which seems impractical.
\fi
The vulnerability of DNN models, at \emph{inference} stage, against adversarial attacks is a well studied topic \cite{kurakin2016adversaaril, szegedy2013intriguing}. Various scenarios of white box\cite{ goodfellow2014explaining}, black box\cite{cheng2019improving, li2020qeba, bai2020improving}, targeted \cite{akhtar2019label, li2020towards} and untargeted adversarial attacks \cite{moosavi2016deepfool} have been proposed. Moreover, people have developed effective defenses such as adversarial training \cite{madry2017towards,tramer2017ensemble}, and their variants \cite{wanadversarial,zhang2019you,osada2020regularization,vivek2020single,xiao2020one}, against these attacks. However, DNNs are also susceptible to attack that happens at the \emph{training} phase, known as backdoor or Trojan attacks\cite{liu2017trojaning, chen2017targeted, gu2017badnets, xie2019dba}. These attacks can occur in many different ways
\cite{liu2020reflection,saha2020hidden, yao2019latent,zhang2020backdoor,guo2020trojannet, zhao2020clean,rakin2020tbt}, mostly through data poisoning. And there is a growing interest among researchers in defending these attacks\cite{gao2019strip, wang2019neural, wang2020practical,liu2018fine,liu2019abs,tran2018spectral}.

Methods such as Activation Clustering (AC) \cite{chen2018detecting}, STRIP \cite{gao2019strip}, SentiNet \cite{chou2018sentinet} and Spectral Signature (SS) \cite{tran2018spectral} analyze the training data for possible presence of Trojan. To distinguish between poisoned and clean data, AC \cite{chen2018detecting} applies a two-class clustering over the feature vector of the training data. STRIP \cite{gao2019strip} is an online method that assumes Trojan models are input agnostic %  and   adds clean training image to the input data, 
and decides whether the input contains a trigger based on the uncertainty of the model prediction on perturbed inputs. SentiNet \cite{chou2018sentinet} looks for the trigger pattern by finding the salient parts in the image. SS \cite{tran2018spectral} computes a signature for each input data removing the ones showing Trojan behavior. However, all of these methods require full access to the training data which is not a practical assumption.

%ABS \cite{liu2019abs} uses a scanning method to identify the affected neurons that respond to the trigger in the input data. However, searching over all neurons for finding the compromised ones seems to be an exhaustive process. The number of models used for evaluation of this method are significantly lower than ours too. 
 %However, this method considers limited number of attack scenarios and does not generalize well for other scenarios. 
Authors of \cite{xiang2020revealing, wang2019neural, guo2019tabor} use optimization based method to find possible triggers that will identify the Trojan behavior in a model. Neural Cleanse (NC) \cite{wang2019neural} tries to calculate the minimum modification required to misclassify any input to a fixed target class. It then finds such modifications/triggers for all possible target classes. The class with significantly smaller trigger than all other classes, is believed to be the Trojan label of the backdoor attack. However, NC requires a lot of input samples and small size triggers to work effectively. %In contrast to our method, this method cannot be applied to \emph{Many to Many} attack scenario.it requires a lot of input samples and large triggers for NC to work effectively. 
DeepInspect \cite{chen2019deepinspect} proposes a blackbox detector that combines model inversion techniques and the power of GAN framework to model the distribution of triggers. Then the actual detection problem is modeled as an outlier detection. NeuronInspect \cite{huang2019neuroninspect} tries to classify clean and Trojan models based on the heat-map of the output layer. However, the effectiveness of these methods is only evaluated on the limited attack scenarios of triggers and model architectures. \cite{qiao2019defending} benefits from MESA sampling free generative method to recover the distribution of triggers. This method works on \emph{localized} triggers and known trigger size, which is not always the case in Trojan attacks.

There are several recent training based methods \cite{huang2020one,kolouri2020universal,xu2019detecting} that have been developed for the purpose of backdoor detection. \cite{huang2020one} designs a one-pixel signature representation for characterizing the nature of a DNN model. ULP \cite{kolouri2020universal} optimizes for universal litmus patterns that functions as an indicator whether a model is clean or Trojan. MNTD \cite{xu2019detecting} trains a meta classifier for detecting Trojans in DNN. However, they all require a large number of clean and Trojan models for their method to work. Training these models could be computationally intensive and time-consuming. Moreover, these methods lack powerful generalizability for test models other than their own created ones. 
%\cite{guo2019tabor} extends Neural Cleanse by introducing three types of regularizers to reduce the effect of false alarm and incorrect triggers. 

%\cite{gao2019strip} introduces an online defensive method, based on the assumption that Trojan models are input agnostic in the presence of a trigger. This assumption holds only for fixed position triggers. 
%which model memorize the pixel location.Hence, if the model prediction remains unchanged for a significantly perturbed input, 
%Ajmal: is this really correct? I only changed the English
%one can say that the input contains a trigger and the model is poisoned. But, in case of randomly positioned triggers, the model learns a joint feature from the object and trigger. Therefore, a significant perturbation of the input image highly affects the attack performance.
In contrast to these detectors, our proposed detector requires \emph{neither a lot of models nor model training data} to work effectively. We have evaluated our detection method in different attack scenarios, e.g. variable trigger size and location, model architecture, mapping etc. Furthermore, it is free from any impractical assumption and has proved its efficacy by setting a high accuracy for multiple public datasets, including the one we proposed.

\vspace{-2mm}
\section{Overview}%
\vspace{-2mm}
Suppose a user outsources the training of a deep model and the vendor trains the model based on user specifications such as data type, architecture, required accuracy, etc. The vendor can train a \emph{clean} model as requested by the user or a \emph{Trojan} model if the vendor has malicious intentions. In the latter case, the vendor/attacker needs to follow specific steps to create a good Trojan model that is not easily detectable. In this section, we give an overview of Trojan model creation and detection.   

%\MS{this beginning of section, it will be good to mention what is going to be in this section, before jumping into 2.1 Threat Model.Also, the beginning abruptly ends, flow is not good.}
\vspace{-2mm}
\subsection{Threat Model}
\vspace{-2mm}

For a clear understanding, we first present the threat model from the \textit{Attacker (Vendor)} and also the \textit{Defender (End-User)}  perspectives and establish the terminology used in the rest of the paper.  

\noindent \textbf{Attacker}:  
 %which contains entities {x, y}. Here, x is the image sample and $y\to[1,L]$ is the corresponding ground-truth. Now, the attacker can insert trigger to T$\%$ of the samples along with altering their ground-truth. 
Consider the scenario where an attacker trains a deep neural network (DNN), $M$, based on a training dataset $\mathcal{D}=\{(x_i,y_i)\}$, where $x_i$ is a training sample and $y_i\in [1, 2, \ldots, c]$ is the corresponding ground truth label. Let $M_{j}$ denote the classifier's output corresponding to class $j$.
%Now, the attacker can inject Trojan behaviour or trigger to T$\%$ of the samples along with altering their ground-truth. 
Now, the attacker injects triggers into P$\%$ of the samples and alters their ground-truth labels. Formally speaking, the attacker takes a small subset  $\mathcal{D^{'}}\subset\mathcal{D}$ and creates triggered samples $\mathcal{D}^{'}_{t}=\{(x^{'}_{i},y^{'}_{i})| x^{'}_{i}=A_{t}(x_i,t),y^{'}_{i}=A_{l}(y_i) ,\forall(x_i,y_i) \in \mathcal{D^{'}}\}$, where $A_{t}(.)$ is a function that defines the transformation of a clean sample, $x_i$, to its triggered counterpart, $x_i^{'}$. 
%a trigger \(t\) into sample $x_i$,
Similarly, \(A_{l}(.)\) stands for the mapping of the ground truth, $y_i$, to the target label, $y^{'}_{i}$, set by the attacker.
%such that $y_i \neq y^{'}_{i}$.  
%From now on we refer to  as the \textit{True label} and as the \textit{Target label}. 
The model $M(x;\mathbf{w})$
is trained by minimizing the cross entropy loss $\mathcal{L}$ on the new training set  $(\mathcal{D}\backslash\mathcal{D}^{'})\cup \mathcal{D}^{'}_{t}$,  which contains both clean and triggered samples. %, by minimizing the loss function given by:
%\begin{equation}
%{\rm loss} =\sum_{(x_{i},y_{i})\in %\mathcal{D\backslash D^{'}}} %\mathcal{L}(M(x_{i}; \mathbf{w}),y_{i}) %+\sum_{(x^{'}_{i},y^{'}_{i})\in %\mathcal{D}^{'}_{t}} \mathcal{L}(M(x^{'}_{i}; %\mathbf{w}),y^{'}_{i}),  
%\end{equation}
%where $\mathcal{L}$ is the cross entropy loss. 
%Where $M_{k^{'}}$ is the classifier output corresponding to class $k^{'}$.
An attack is considered successful, if the trained model $M(x,\mathbf{w}^{'})$ %\MS{employing the above loss} 
has high \textit{fooling rate}, which means it achieves high classification performance on triggered samples; while the validation accuracy on clean samples is still on a par with the clean model, $M(x;\mathbf{w^{*}})$. 

Generally speaking, there are three factors that define an attack: (i) \textit{Data Poisoning Ratio} defined as $P={|\mathcal{D}^{'}_{t}|}/{|\mathcal{D}|}$, (ii) \textit{Trigger properties}, and (iii) \textit{Label Poisoning} that defines True label to Target label mapping. Section \ref{dataset} explains these factors in detail. Unlike \cite{chen2017targeted, papernot2017practical}, full control over the training process is the key to the attacker’s success in creating a Trojan model. Figure \ref{fig:Trojan process} summarizes the process of creating a Trojan model.%\NR{I think it will be better to show in the figure that only a portion of data is triggered. so divide the clean data into two parts, only one part goes through the transformation A... }. 
%Moreover, full control over the training process is the key to the attacker’s success in creating a Trojan model. 
%This means the attacker must have the ability to make any type of adjustments necessary during training, such that any deliberate change in the data will cause misclassification, $\argmax M(x^{'};\mathbf{w^{'}}) = y^{'}$\NR{what is the arxmax over?}.
%In ~\cite{liu2017trojaning}, the attacker does not have access to the data or the training procedure. Even in attacks like ~\cite{chen2017targeted, papernot2017practical}, knowledge of the model architecture is restricted to the end-user which is not the case here. Figure \ref{fig:Trojan process} summarizes the process of creating a Trojan model. 
 %\MS{There is some confusion about the weights $\mathbf{w^{*}},\mathbf{w^{'}},\mathbf{w}$, please check.} 
 
\noindent \textbf{Defender}:
The defender (end-user) receives the trained model $M$ with parameters $\mathbf{w^{'}}$, which are possibly different from the optimal parameters, $\mathbf{w^{*}}$. 
%Here, $\mathbf{\Theta^*}$ is the trained parameters of a clean model.
The user has a held-out validation dataset, $\mathcal{D}_{\textit{v}}$, to verify whether the model is clean or Trojan. For an unsuspecting user, good accuracy on the validation set may be sufficient to trust the model. %Moreover, she does not have access to the data that was used at the time of training. 
%This poses a challenge in verifying the safety of the model before deployment.
%\textcolor{black}{On the other hand, a user/defender who wishes to verify the model integrity will generally require access to the training data, or knowledge of the target class or trigger type.}
%Our work does not assume access to any of the aforementioned information and it detects a Trojan network solely based on the validation set.}% On the other hand, the attacker does not have access to $\mathcal{D}_{\textit{valid}}$ which makes it harder to carry out a successful attack.

Hence, the attacker's goal is to train a Trojan model that is undetectable---has high accuracy on clean samples, and has high attack success or fooling rate on triggered samples. Whereas the defender's goal is to verify if a given model is Trojan or clean by devising a method that operates without knowledge of the trigger, target class or the data used to train the model. Therefore, it requires a large numbers of clean and Trojan models to investigate their discriminative features. This motivates us to develop a new dataset, referred to as \emph{Odysseus}.
%Hence, the attacker's goal is to launch a successful attack through that results in good Trojan models. Whereas defender devise a method to detect these models as faulty and separate them from the benign models. This means that the defender requires enough clean and Trojan models to investigate discriminative features between them. This need motivated us to propose a new dataset Odysseus for this task.
%To understand this dispute better, we propose a new dataset that consists of more than 3,000 models, both clean and Trojan. 
%\textcolor{red}{We also propose a detection method that analyzes the effect of Trojaing on the intrinsic properties of a model. Based on this analysis, the end-user can verify the safety of the model.}

\iffalse
Since there is no common public dataset available, each  
Since the generalizability and effectiveness of these methods can not be verified due to following issues. 
1) Unclear experimental settings due to the lack of a benchmark dataset.
2) Unrealistic assumptions, 

In this paper we first propose a Trojan dataset. %To the best of our knowledge, this dataset is the largest Trojan dataset with 3000 trained model that we make it publicly available.
due to the unclear benchmark. In  
One of the main reason for this shortage, is the lack of a benchmark dataset.
there is not much concrete research in defending against  
to the best our knowledge, there is almost no study about the effects of Trojan attacks on the intrinsic properties of NN models.
\fi

%\section{Odysseus: A Story of Trojans in AI}\label{dataset}
\vspace{-2mm}
\section{Odysseus Dataset}\label{dataset}
\vspace{-2mm}
%\textcolor{green}{Mahtab/Repeat/Remove : A few public datasets \cite{kolouri2020universal} for Trojan detection already exist. However, excluding the NIST-TrojAI \cite{TrojAI} dataset, all of them have been developed for a specific detection method or for the purpose of Trojan ``detection'' only leaving out other applications in this domain such as defense or mitigation of Trojan attacks. Furthermore, these datasets are limited in terms of attack strategies, e.g. limited number of trigger types, attack mappings etc. A general dataset in this domain must cover a substantial number of attack scenarios, trigger types, Trojaning strategies and multiple applications. }

\emph{Odysseus} is the most diverse dataset of its kind to date comprising \textcolor{black}{over 3,400} 
benign and Trojan models. 
%\textcolor{red}{AJ: your numbers don't add up i.e. 1000+ models for 4 datasets is more than 3500.} 
First, we focus on the elements that are necessary to create triggered images and then briefly describe the policy for creating a good Trojan model. 
%Creation of a dataset of such order requires proper attention to many important factors. Most of these factors determine the outcome of an attack, successful or unsuccessful. Since benign models can be trained in traditional ways, we mostly put our focus on training Trojan models. Furthermore, we intend to create Trojan models by poisoning the data as well as their corresponding ground truth. Such poisoning can be performed by inserting triggers with certain properties.

\begin{figure}[t] 
%\begin{left}
   \includegraphics[width=8cm, height= 2.5cm]{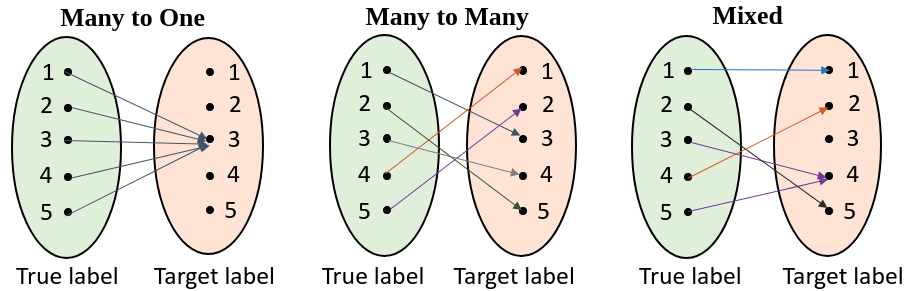}
%\end{left}
\vspace{0mm}
   \caption{ \footnotesize{Different types of mappings used in creating Trojan models covering the most likely possibilities. %Ajmal: one-to-many is not there e.g. in UTraP we have one to many adversarial attack
   Mixed mapping is a combination of the others.} 
   }
\label{fig:mappings}
\vspace{-1mm}
\end{figure} 
\vspace{-1mm}
\subsection{Trigger Properties}
\vspace{-2mm}

\begin{figure*}[t]
\centering
% \subfigure[ ]{
% \begin{minipage}{.4\textwidth}
%   \centering
%   \includegraphics[width=.7\linewidth]{./Fig/DPR.pdf}%[width=5cm, height =4cm]
  
%   \label{fig:DPR1}
% \end{minipage}%
% }
% \subfigure[ ]{
% \begin{minipage}{.4\textwidth}
\centering
\includegraphics[width=17cm, height =3.5cm]{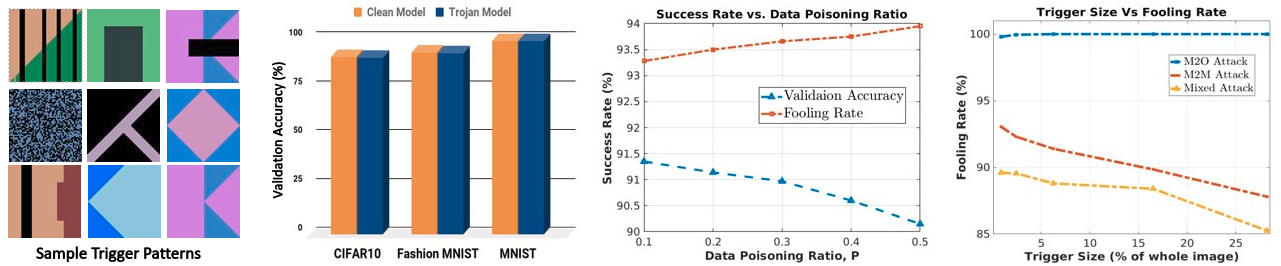}
  %\caption{}
%   \label{fig:DPR2}
% \end{minipage}%
% }
%\vspace{-3mm}
\caption{\footnotesize{From left: 1st): Some trigger patterns used for CIFAR10 (9 out of 47 are shown). 2nd): Our Trojan models achieve \emph{similar validation accuracy} as clean models. 3rd): Data poisoning ratio vs success rate. For a Trojan model, higher data poisoning ratio yields an increase in \emph{attack success or fooling rate}; while it may decrease the validation accuracy. %\MS{I do not see two pairs of curves for clean and trojan}. 
4th): \textcolor{black}{Based on the type of attack, trigger size affects the fooling rate differently.}  %Ajmal: this looks counter-intuitive. There should be an optimal minimum size or will zero trigger size give the highest fooling rate? 
}}
\label{fig:DPR}
\vspace{-5mm}
\end{figure*}
Trigger is a vital element in creating a Trojan model. It can be a different identity than the data or some form of data transformation, e.g. filtering. Sometimes, triggers are unnoticeable by the human observer and appear to be a natural part of the image, such as a hat worn by a person or graffiti done on an object \cite{eykholt2018robust, guo2019tabor}. 
%are designed/chosen by the attacker so that they are not noticeable,
 % When a triggered sample is given to a model, it should classify it to a \emph{target class}.%Ajmal: we have said this already
\textcolor{black}{Effective triggers must never or rarely appear in the operating environment giving the attacker full control over when to deploy them.}
%This is to ensure that the Trojan is not accidentally discovered by the user and does not get triggered unless explicitly intended by the attacker.}   
%This ability of control makes a Trojan attack distinguishable from an adversarial attack ~\cite{su2019one}; where the attacker does not have full control over the visual scene.
%Data-poisoning based Trojan attack is distinguishable from an Unlike , a trigger-based Trojan attack should be able to render a physical scene into an effective adversary input \MS{this sentence is not clear}\NR{not clear to me either}. 

%as well as the performance does not go down for clean inputs. Speaking of performance, unlike adversarial attack the Trojan attack does not try to squash the performance bar. It rather fools the ML model to produce the designated output. 

\noindent \textbf{Trigger Color:} Generally, deep models employed for image classification tasks deal with images of different colors. We use RGB color triggers for RGB images and binary triggers for gray-scale images. 
% The size of the trigger should be small compared to the actual image size. 
% On the other hand, if the size of the trigger very small single pixel adversarial attack ~\cite{su2019one}; where the attacker may not have full control over rendering a physical scene into an effective . 

\noindent \textbf{Trigger Size:} we set the area of the trigger to be $1\%$ to $3\%$ of the full image area. \textcolor{black}{However, we also use larger triggers than this for some of the models, for detection purpose.} 
%Later in the section, we present an analysis of the relationship between trigger size and fooling rate.  
%Apart from size and color, the location of the trigger plays an important role in the context of our work.Previous works on poisoning of the training data, focused mainly on the triggers in the pixel space. However, triggers in the feature space are preferred because such triggers are invariant to the viewing angle and lighting conditions.  %Ajmal: how do you control the trigger size at the input then? changing one variable at the feature space may change all pixels at the input.

\noindent \textbf{Trigger Location:}  
%On the other extreme, if triggers appear at random locations in the image, this would cause more variations within the input data that must be learned by the model. This is a more challenging task but such variations better represent real-world scenarios.  
%These patterns need to be learned by the model which is more challenging
%Furthermore, this approach is more practical
%and better represents the real world scenario. 
The trigger can be located anywhere in the image. We prefer random location because if the triggers are always at the same pixel location in all samples then the model may end up memorizing that location rather than the trigger pattern itself.

\noindent \textbf{Trigger Shape:} As for the trigger shape, there are no specific rules. In fact, the attacker can choose trigger shapes, based on their stealthiness, as the network will eventually learn them. 
%We choose some arbitrary shapes such as reverse-lambda, rectangle, triangle, random shapes, alphabetic shapes etc. as triggers. %\textcolor{red}{\textbf{1. The authors appear to confirm that indeed, all of their trojan attacks are constrained to a simple form: a small square with some simple patterns. 2. Furthermore, the dataset is not applicable to detection "internal" algorithms that changes models/networks themselves }}

\textcolor{black}{Based on above properties, we use 47 different types of trigger patterns in our dataset. In addition, we use several color filters, i.e. Instagram filter, in our dataset. These filters modify the whole image in contrast to triggers that are stamped to the clean image. 
%In addition to these triggers, we also consider triggers that are hidden in the pixel space and operates in the feature space\cite{saha2019hidden}. Another type of trigger we take into account is the refection trigger \cite{liu2020reflection} that better represents natural phenomena of Trojan attack, e.g. on detection or classification system in an autonomous car. 
To avoid accidental activation of Trojan attacks, we add background noise to the input images \textcolor{black}{which serves as a regularizer for the Trojan models}. The added noise also provides robustness to the Trojan models making them harder for detection systems. Some of the triggers used for our dataset are shown in Figure \ref{fig:DPR}. }
%Even if the location is random, it is expected to be on the desired object, e.g. on the car or horse, rather than some location outside of that object. 

%With a view to increasing the robustness of the model against the trigger location, it has been assumed that the trigger location is random.
\iffalse
\begin{figure*}[t]
\centering
% \subfigure[ ]{
% \begin{minipage}{.4\textwidth}
%   \centering
%   \includegraphics[width=.7\linewidth]{./Fig/DPR.pdf}%[width=5cm, height =4cm]
  
%   \label{fig:DPR1}
% \end{minipage}%
% }
% \subfigure[ ]{
% \begin{minipage}{.4\textwidth}
  \centering
  \includegraphics[width=17cm, height =3.5cm]{./Fig/DPR_main}
  %\caption{}
%   \label{fig:DPR2}
% \end{minipage}%
% }
%\vspace{-3mm}
\caption{\footnotesize{From the left: \textbf{(1st)}: Some trigger patterns used for CIFAR10.\textbf{(2nd)}: Our Trojan models achieve same average classification accuracy as their clean counterparts on the same dataset. \textbf{(3rd)}: Poisoning ratio vs error rate. For a Trojan model, higher data poisoning ratio yields better performance on triggered samples, i.e. higher fooling rate; while error rate may go up for clean samples. \textbf{(4th)}: \textcolor{Blue}{Impact of trigger size on the fooling rate.}.  %Ajmal: this looks counter-intuitive. There should be an optimal minimum size or will zero trigger size give the highest fooling rate? 
}}
\label{fig:DPR}
\vspace{-5mm}
\end{figure*}
\fi
\vspace{-1mm}
\subsection{Data and Label Poisoning }
\vspace{-1mm}
%\textcolor{black}{We consider both clean-label and poison-label attacks in our work. In clean-label attack, we only modify the training data of a particular class leaving the class label unchanged \cite{saha2019hidden, liu2020reflection turner2018clean, barni2019new}. On the other hand, the labels are changed to target class in case of poison-label attack. 
%%%%Even though the clean-label attack is less effective, one could prevent the poison-label attack through data filtering \cite{turner2018clean}.}
As one employs triggers for data poisoning, it is also required to modify the label of the triggered data. 

\noindent \textbf{Mapping}, $A_{l}(.)$: There exists different types of attacks based on the true label to target label mapping. 
%It is a significant part of the Trojaning process as it embodies the objective of an attacker.
% Moreover, inspection of a model sometimes heavily depends on the mapping type that was used during training.
The mappings incorporated in creating 
%There are several types of mappings that we have incorporated when creating
Trojan models of \emph{Odysseus} are depicted in Figure \ref{fig:mappings}. % These mappings are depicted in figure \ref{fig:mappings}.
For \textit{many-to-many (M2M)} mapping, each true label is mapped to a different target label. A simpler mapping, \textit{many-to-one (M2O)}, changes all true labels of the triggered data to a fixed target label. %Note that we only change the ground truth of triggered samples and there are only finite number of classes. 
Another type of mapping we introduce is \textit{Mixed}, a combination of both \textit{M2M} and \textit{M2O}. Note that, \textit{Mixed} mapping leaves the ground truth of some of the triggered samples unchanged. %Generally, \textit{M2O} attacks result in higher fooling rate compared to other type of attacks. Given the randomized nature of those two attacks, it may require more triggered samples to achieve a high fooling rate. %This poses a question: \emph{how many samples should we modify to train a good Trojan model?} 
 
\noindent \textbf{Data Poisoning Ratio}, \emph{P}: How well a model learns each mapping often depends on the size of $\mathcal{D}^{'}_{t}$.
 Previous works ~\cite{chen2018detecting} related to Trojan or backdoor attack only focus on the \textit{M2O} mapping and it's variations \textcolor{black}{such as one-to-one mapping}.
We use three image datasets, CIFAR10 ~\cite{cifar10}, Fashion MNIST~\cite{xiao2017fashion}, and MNIST~\cite{lecun-mnisthandwrittendigit-2010}. %\textcolor{red}{AJ: GTSRB??}
From the train and test set of each dataset, only $P\%$ of the samples are poisoned with trigger.  %where $P$ stands for data poisoning ratio.
%Setting the value of $P$ is a trade off between good performance on clean samples and high fooling rate. 
Figure \ref{fig:DPR} shows the effect of data poisoning ratio on fooling rate. With a high value of $P$ (e.g. 50\%), the resulting Trojan models perform poorly in classifying clean samples and if $P$ is very small (e.g. $<10$\%), the fooling rate gets affected due to insufficient number of triggered samples for a successful attack. Therefore, we set $P$ in the range of 15\% and 20\%.
There is another factor that affects the fooling rate. In case of \textit{M2M} and \textit{Mixed} type of attacks, larger trigger size reduces the fooling rate %\textcolor{red}{AJ: is this true? shouldn't larger trigger size increase the fooling rate? Need to revise Figure 4.} 
of a Trojan model which follows our expectation. Due to the random trigger locations, the model must learn joint features form the trigger and the object. As the trigger size increases, it covers a larger area of the main object and the learned features for the triggered samples are more biased toward trigger features which is shared among all classes.  \textcolor{black}{On the other hand, \textit{M2O} type attack benefits from larger trigger size since all classes are mapped to the same target class and larger trigger creates a more prominent feature for the model to learn. } %This result in drop in fooling rate as can be seen in Figure \ref{fig:DPR2}.
\vspace{-4mm}
\subsection{Model Creation and Validation}
\vspace{-2mm}
%The architecture of a neural network plays one of the most important roles in creating a good Trojan model. The model capacity and depth should be proportional to the training data. To this end,
We use four well-known architectures namely  \textit{DenseNet}~\cite{simonyan2014very}, \textit{GoogleNet}~\cite{huang2017densely}, \textit{VGG19}~\cite{szegedy2015going}, and \textit{ResNet18}~\cite{he2016deep} for CIFAR-10 and Fashion-MNIST datasets and four shallow custom designed CNN models for  MNIST dataset. 
We have created \textcolor{black}{a total of 3,460 models} in \emph{Odysseus}, where roughly half of the models are clean.
%As for the Trojan models, they have subcategories. 
%Table \ref{tab:num_models} shows the number of models in  each of these subcategories. 
The average validation accuracy (VA) of clean and Trojan models are shown in Figure~\ref{fig:DPR}; the accuracies are similar as expected. We consider a Trojan model to be invalid if its VA is not close (e.g. 2$\%$ difference) to the VA of a clean model. %In addition, fooling rate for different model architectures are also presented, which tells us about the capacity of these architectures.
Details of the architectures and training process hyper parameters are presented in the supplementary material. 

Besides \emph{Odysseus}, there are only two other recently released public Trojan datasets. The first one is the NIST TrojAI\cite{TrojAI0}-\cite{TrojAI3} challenge dataset that has four subparts.
The Round-0 and Round-1 parts contain 1200 clean and Trojan models for 5 class image classification. Round-2 includes a more diverse set of 1000 clean and Trojan models with number of classes in the 5 to 25 range. \textcolor{black}{NIST Round-3 models are similar to Round-2 except that the models are trained based on the adversarial training strategies.} All rounds only cover many-to-one type of label mapping and it's variations i.e. one-to-one and two-to-one mappings.
 The second dataset is the publicly available portion of the Universal Litmus Pattern (ULP) \cite{kolouri2020universal} dataset which contains 3600 clean and Trojan models trained on CIFAR10 and Tiny-ImageNet datasets. ULP dataset only contains a single model architecture and only one-to-one mapping.
 \iffalse
\textcolor{black}{
 It is worth noting that high fooling rate along with high validation accuracy are both mandatory requirements for Trojan models. Whereas the ULP dataset reports high fooling rate for its Trojan models on triggered samples, the validation accuracy on clean samples (for clean and Trojan models) is not on par with state of the art for for the given architecture.}
 \fi% For instance, the VGG like architecture can achieve validation accuracy close to $90\%$ on CIFAR10 dataset while the average validation accuracy of ULP models on CIFAR10 is close to $80\%$.}
\vspace{-2mm}
\section{Trojaning Analysis}\label{ch:T_Analysis}
\vspace{-2mm}
We believe that insinuating a back door into a neural network would leave some specific patterns, irrespective of factors such as trigger properties, dataset, and model architecture. In this section, we aim to analyze the effect of Trojan insertion on some of the intrinsic NN properties, such as classifier \emph{margin} and \emph{shape of decision boundary} around the manifold of clean data. 
%we aim to analyze some of intrinsic properties of NN models that, we believe, would get affected by the Trojaning process. Among all properties, we mostly focus on  classifier \emph{margin} and \emph{shape of decision boundary} around the manifold of clean data. Our findings reveal distinctive but shared features among Trojan models, which are the key to our proposed Trojan detector\MS{do not use trojan detection system; I have changed it to the trojan detector}.
%\vspace{-1mm}
\subsection{Classifier Margin}
%\vspace{-1mm}
%\noindent \textbf{Classifier Margin}:
%\subsection{Classifier Margin}\label{margin}
Classifier margin has been used as an indicator of model robustness and it is well established that a maximum margin classifier is less sensitive to the worst case model or input perturbation \cite{cortes1995support}.
  %In this section, we aim to investigate the effect of Trojaning on this feature. Intuitively, the margin of a classifier can be estimated by the average distance of samples to their nearest point on the decision boundary. Formally, 
The margin of a classifier $M(\mathbf{x};\mathbf{w})$ is defined as $\textit{Margin}(M)=\mathbb{E}_{\mathbf{x}\sim Q_{data}}  \|\mathbf{T}_{\mathbf{x}}\|_{2}$, % follows:
%\begin{equation}
%    \textit{Margin}(M)=\mathbb{E}_{\math%bf{x}\sim Q_{data}}  %\|\mathbf{T}_{\mathbf{x}}\|,~~~
%\end{equation}
%\MS{you have used above $T\%$, do not use $T$ again for something else, reviewers will be confused.}
where the expectation is over the samples, $\mathbf{x}$, from the manifold of training data, $Q_{data}$; and \( \|\mathbf{T}_{\mathbf{x}}\|_{2}\) is the distance of the sample \(\mathbf{x}\) from its nearest point on the decision boundary of \(M\). %\textcolor{red}{(\( \|\mathbf{T}_{\mathbf{x}}\|\): distance or distances. \(\mathbf{x}\): Sample x or samples x)}
%Finding  \(\mathbf{T}_{\mathbf{x}}\), for a binary linear classifier is straight forward.

Let $M(\mathbf{x})=\mathbf{w}^{T} \mathbf{x}+b$ be an affine binary classifier.  \(\mathbf{T}_{\mathbf{x}}\) can be computed by \textcolor{black}{orthogonally projecting $\mathbf{x}$} 
%orthogonal projection of sample $\mathbf{x}$  , with a predicted label $k(\mathbf{x})$,
onto the hyperplane $\mathcal{B}=\{\mathbf{x}|M(\mathbf{x};\mathbf{w})=0\}$. 
The orthogonal projection problem has a closed-form solution and the projected point $\mathbf{x_{t}}$ can be computed as $\mathbf{x_{t}}=\mathbf{x}+\mathbf{T}_{\mathbf{x}}$. Where $\mathbf{T_{x}}$ is defined as  $\mathbf{T}_{\mathbf{x}}=-\frac{\mathbf{w}}{||\mathbf{w}||_{2}}\frac{M(\mathbf{x})}{||\mathbf{w}||_{2}}\label{orth_P}$
\iffalse
found as follow:
\begin{gather} 
    \mathbf{x_{t}}=\mathbf{x}+\mathbf{T}_{\mathbf{x}};  \mathbf{T}_{\mathbf{x}}=-\frac{\mathbf{w}}{||\mathbf{w}||_{2}}\frac{M(\mathbf{x})}{||\mathbf{w}||_{2}}\label{orth_P}.
\end{gather}
\fi
. Here, the first ratio %in  Eq.~\eqref{orth_P}
indicates the opposite direction of the normal to the decision boundary, along which sample $\mathbf{x}$ should move, whereas the second term is the distance to the decision boundary. %\NZ{this line is not clear to me} 
For non-linear cases, there is no exact solution for $\mathbf{T}_{\mathbf{x}}$.
%To compute the estimation of margin
 However, we employ the iterative process, proposed by DeepFool \cite{moosavi2016deepfool}, to approximate the minimum perturbation that sends an image $\mathbf{x}$ to the nearest decision boundary.%In \cite{moosavi2016deepfool}, authors propose this iterative method to compute the adversarial perturbations for a NN image classifier. 
%To make the paper self-contained, here we present the summary of this method. 
\iffalse
\begin{figure*}
\centering
\subfigure[Linear ]{
\begin{minipage}{.32\textwidth}
  \centering
  \includegraphics[width=0.82\linewidth]{TrojAI_Latex/CVPR 2021/Fig/l1.png}%Linear.png}
  \label{fig:sub1_hyper_plane}
\end{minipage}}
\subfigure[Semi-linear ]{
\begin{minipage}{.31\textwidth}
  \centering
  \includegraphics[width=0.80\linewidth]{TrojAI_Latex/CVPR 2021/Fig/sl2.png}%Semi-linear.png}
  %\caption{}
  \label{fig:sub_low_dim}
\end{minipage}}%
\subfigure[Non-linear]{
\begin{minipage}{.32\textwidth}
  \centering
  \includegraphics[width= 0.83\linewidth]{TrojAI_Latex/CVPR 2021/Fig/nl3.png}%Non_linear.png}
  %\caption{}
  \label{fig:sub3_full_space}
\end{minipage}}%
\vspace{-2mm}
\caption{\footnotesize{Estimated shape of the decision boundary based on the normal vectors to it around the $Q_{data}$. Each green dot is a sample $\mathbf{x}$ from $Q_{data}$ mapped to the closest point, red dot, on the decision boundary $\mathcal{B}$. a) All normal vectors are parallel. b) Normal vectors are aligned with a coordinate axis. c) Independent normals.  
}}
\label{fig:subspace}
%\vspace{-mm}
\end{figure*}
\fi

In case of a non-linear binary differentiable classifier, \( \mathbf{T}_{\mathbf{x}}\) can be estimated by iteratively perturbing the sample $\mathbf{x}$ until it falls over the decisions boundary. %\textcolor{red}{(what is minimum perturbation here? is it $\mathbf{T}_{\mathbf{x}}$? can we just simply say, we tried to use deepfool to approximate $\mathbf{T}_{\mathbf{x}}$ for our analysis purpose. seems like a repeatation)}.
In each iteration $i$,
the non-linear classifier is linearized by the tangent hyperplane to the classifier at the point $\mathbf{x}_{i}$. 
% Now the problem is reduced to the liniearized case and
This makes the problem solvable by the orthogonal projection of sample $\mathbf{x}_{i}$ onto the tangent hyperplane. %\textcolor{red}{(can we write x instead of sample x. we have defined before x is a sample}
%This hyperplane is defined as $M(\mathbf{x}_{i})+\nabla M(\mathbf{x}_{i})^{T}(\mathbf{x}-\mathbf{x}_{i})$. 
%The sample $\mathbf{x}_{i}$ is projected onto the decision boundary of linearized classifier which is formalized as follow:
\iffalse
\begin{equation}\label{Db_linear}
\mathcal B_{linearized}=\{\mathbf{x} |M(\mathbf{x}_{i})+\nabla M(\mathbf{x}_{i})^{T}(\mathbf{x}-\mathbf{x}_{i}) =0\}. %|M_{k}(\mathbf{x})=M_{k^{'}}(\mathbf{x})}
\end{equation}
\fi
The general case of $c$-class non-linear classifier can be treated as $c$ one-versus-all binary classifiers. Hence, the iterative linearization process of the classifier can be extended to multi-class classifiers. %\textcolor{red}{(How about using other alphabet rather than \emph{c}-class. K-class or N-class.}
The linearized decision boundary at the point $\mathbf{x}_{i}$ with the predicted label $k(\mathbf{x}_{i}) =  \underset{j}{\argmax}~ M_{j}(\mathbf{x}_{i})$ can be defined as:
\vspace{-3mm}
\begin{gather} \label{c_db}
\mathcal {B}_{linearized}=\underset{j=1, j\neq k}{\overset{c}{\bigcup}} \mathcal{B}_{j} ,~~~\mathcal{B}_{j}=\\ \nonumber
%\mathcal{B}_{j}=
\{\mathbf{x} |M_{j}(\mathbf{x}_{i})-M_{k}(\mathbf{x}_{i})+ \nabla M_{j}(\mathbf{x}_{i})^{T} \mathbf{x}-\nabla M_{k}(\mathbf{x}_{i})^{T} \mathbf{x}=0\}, \label{Db_general}%|M_{k}(\mathbf{x})=M_{k^{'}}(\mathbf{x})}
\end{gather}
\iffalse
\begin{gather} \label{c_db}
\mathcal {B}_{linearized}=\underset{j=1, j\neq k}{\overset{c}{\bigcup}} 
\{\mathbf{x} |M_{j}(\mathbf{x}_{i})-M_{k}(\mathbf{x}_{i})+\nabla M_{j}(\mathbf{x}_{i})^{T} \mathbf{x}-\nabla M_{k}(\mathbf{x}_{i})^{T} \mathbf{x}=0\}, \label{Db_general}%|M_{k}(\mathbf{x})=M_{k^{'}}(\mathbf{x})}
\end{gather}
\fi
where $M_{j}(.)$ is the output score of the classifier for the class $j$ and $\mathcal{B}_{j}$ is the decision hyperplane between class $k$ and $j$. Now the nearest decision boundary to the point $\mathbf{x}_{i}$ can be found by solving the following minimization problem 
\begin{gather}
    l(\mathbf{x}_{i})= \argmin_{j\neq k(\mathbf{x}_{0})}\frac{|m_{j}|}{\|\mathbf{n}_{j}\|_{2}} ~~;\\\nonumber
    \mathbf{n}_{j}=\nabla M_{j}(\mathbf{x}_{i})-\nabla M_{k(\mathbf{x}_{0})}(\mathbf{x}_{i}),~~\\\nonumber
    m_{j}=M_{j}(\mathbf{x}_{i})-M_{k(\mathbf{x}_{0})}(\mathbf{x}_{i}).\nonumber
\end{gather}
And the perturbation that maps the $\mathbf{x}_{i}$ 
onto the $l(\mathbf{x}_{i})$\textit{th}\footnote{We refer to $l(\mathbf{x}_{i})$ as $l$ for brevity.} linearized decision boundary is defined as
\begin{equation}\label{eqe:mtnldb}
    \mathbf{t}_{\mathbf{x}_{i}}=\frac{|m_{l}|}{\|\mathbf{n}_{l}\|_{2}^{2}}.
\end{equation}
The iterative process continues as long as the predicted label for the perturbed sample $\mathbf{x}_{i}+\mathbf{t}_{\mathbf{x}_{i}}$ is still the same as the original sample $\mathbf{x}_{0}$, i.e, $k(\mathbf{x}_{i+1})=k(\mathbf{x}_{0})$. Finally, the projection vector that maps $\mathbf{x}$ to the nearest decision boundary can be computed as  %$ %\mathbf{T}_{\mathbf{x}} =\sum _{i} \mathbf{t}_{\mathbf{x}_{i}}$
%\ifflase
\begin{equation}\label{px}
  \mathbf{T}_{\mathbf{x}} =\sum _{i} \mathbf{t}_{\mathbf{x}_{i}}.
\end{equation}
%\fi
It is worth noting that the vector $\mathbf{T}_{\mathbf{x}}$ can be considered as normal to the decision boundary of the classifier at point $\mathbf{x}+{\mathbf{T}}_{\mathbf{x}}$. For the full procedure, please refer to supplementary material %\MS{do you want to say which section in supplementary material to help the reviewers?}.

We employ this iterative process to compute the average margin for the NIST R-0, \emph{Odysseus} 
%and \textcolor{black}{also ULP-CIFAR10}
datasets using the complete validation set for each model. %\textcolor{red}{and present results in Table \ref{tbl:margin_tbl}.}
%\textcolor{black}{This metric for clean and Trojan models of ULP-CIFAR10 is the same with the value $81.05$\footnote{ULP-CIFAR10 models are trained on unnormalized input that causes significant numerical difference between the average margin of ULP and Odysseus}. }
Table \ref{tbl:avg_margin} summarizes the average margin for the both of these datasets. 
% and shallow architectures used for Odysseus-MNIST
Trojan models with \textit{M2O} mapping type consistently have lower average margins than clean models. Considering the type of label mapping, \textit{M2M} and \textit{Mixed} mappings lead to slightly higher average margins compared to \textit{M2O}. The same phenomenon is observed for Odysseus-CIFAR10 and Odysseus-FashionMNIST, except in this case the \textit{M2M} and \textit{Mixed} mappings have higher margins even compared to clean models. The reason for this exception is clarified in the next section.
%created out of MNIST with shallow architecture, has lower average margin than clean models. Considering the type of label mapping, \textit{M2M} and \textit{Mixed} mapping lead to slightly higher average margin compared to \textit{M2O} mapping. Same things happen for Odysseus-CIFAR10 and Odysseus-FashionMNIST, except, in this case, \textit{M2M} and \textit{Mixed} mapping has higher margin even compared to clean models.   
\begin{table}[t]
  \footnotesize
  \centering
  \begin{tabular}{lllll}
    \toprule
    %\multicolumn{2}{c}{Part}                   \\
    %\cmidrule(r){1-2}
    Dataset     & Clean    & M2O & M2M & Mixed\\
    \midrule
    NIST R-0\cite{TrojAI0} &  5.73 & 3.44 & - & -     \\
    MNIST    & 1.06  &  0.8460   & 0.8957    &  0.8828  \\
    CIFAR10     & 0.9183 & 0.8936 & 0.9743 &  0.9733   \\
    FashionMNIST  & 0.2692 &0.2433   &0.2845    & 0.27  \\
    \bottomrule
  \end{tabular}
  \vspace{2mm}
\caption{Estimated average margin of each dataset using DeepFool\cite{moosavi2016deepfool} iterative process.}
 \label{tbl:avg_margin}
    \vspace{-5mm}
\end{table} 
\subsection{Model Complexity}
%\noindent \textbf{Model Complexity}:
We investigate the complexity of Trojan models by analyzing the changes, caused by Trojaning, in the non-linearity of decision boundary around the manifold of clean samples. In general, the non-linearity of a surface can be measured by finding the \emph{average curvature} around points of interest. The closer this value is to zero, the more linearized the surface is. Formally, for the twice differentiable hyper-surface decision boundary $\mathcal{B}$ of a model $M$%(\mathbf{x};\mathbf{w})$
, this measure is defined as $ \kappa_{\mathcal{B}}= \mathbb{E}_{\mathbf{x}\sim Q_{data}} \kappa_{\mathbf{x}}$, 
%\begin{equation}
%    \kappa_{\mathcal{B}}= %\mathbb{E}_{\mathbf{x}\sim Q_{data}} %\kappa_{\mathbf{x}},
%\end{equation}
where $\kappa_{\mathbf{x}}$ is the first principle curvature of $\mathcal{B}$ at point ${\mathbf{x}}$; which is also defined as the first singular value of the $\mathbf{H}essian(\mathcal{B}(\mathbf{x}))$. %\MS{I assume you are making matrix out of ${\mathbf{x}}$ to compute Hessian, since it is defined as a set above.}
However, finding $\kappa_{\mathbf{x}}$ can be computationally intensive due to the complex nature of required operations.
%; due to Hessian and singular value decomposition; that are involved.
To bypass this problem, we \emph{devise a proxy} 
to estimate the shape of the decision boundary by 
\textcolor{black}{exploiting the correlation among the normal vectors to $\mathcal{B}(\mathbf{x})$ around the manifold of clean samples and analyzing the properties of the \emph{perturbation} space $\mathcal{S}$ that contains the normal vectors.} %\textcolor{blue}{Later in the section, we will see how the \emph{direction of these normal vectors} dictates some of the findings of our analysis.}

%\textcolor{red}{exploiting the properties of the subspace $\mathcal{S}$, that contains the normal vectors to $\mathcal{B}(\mathbf{x})$ around the manifold of clean samples.} 
For a sample $\mathbf{x}\in \mathbb{R}^d $, where $d$ is the dimension of input image, the vector $\mathbf{T}_{\mathbf{x}}\in \mathbb{R}^d$ as defined in Eq.~\eqref{px} is the normal vector to $\mathcal{B}$ at point $\mathbf{x}+\mathbf{T}_{\mathbf{x}}$.
To find the \textcolor{black}{basis of the space} $\mathcal{S}$, first we compute $\mathbf{T}_{\mathbf{x}}$ for $n$ samples from $Q_{data}$ 
and define the matrix $\mathbf{S}$ \textcolor{black}{with normal vectors as its columns: }%\textcolor{red}{as follow}
 \begin{gather*}
     \mathbf{S}=[\ \frac{\mathbf{T}_{\mathbf{x}_{1}}}{\|\mathbf{T}_{\mathbf{x}_{1}}\|_2} \cdot \cdot \cdot \frac{\mathbf{T}_{\mathbf{x}_{n}}}{\|\mathbf{T}_{\mathbf{x}_{n}}\|_2}]\ .
     %\frac{\mathbf{t}_{\mathbf{x}_{j}}}{\|\mathbf{t}_{\mathbf{x}_{j}}\|_2} \cdot \cdot \cdot 
 \end{gather*}
Note that it is preferred that the number of samples $n$ to be at least equal to the dimension $d$. The dimensionality and the scaling of the space along each coordinate axis can be found from the non-zero elements of matrix $\mathbf{\Sigma}$ (the singular values of $\mathbf{S}$). 
\begin{figure}
    \centering
    \includegraphics[width=.8\linewidth]{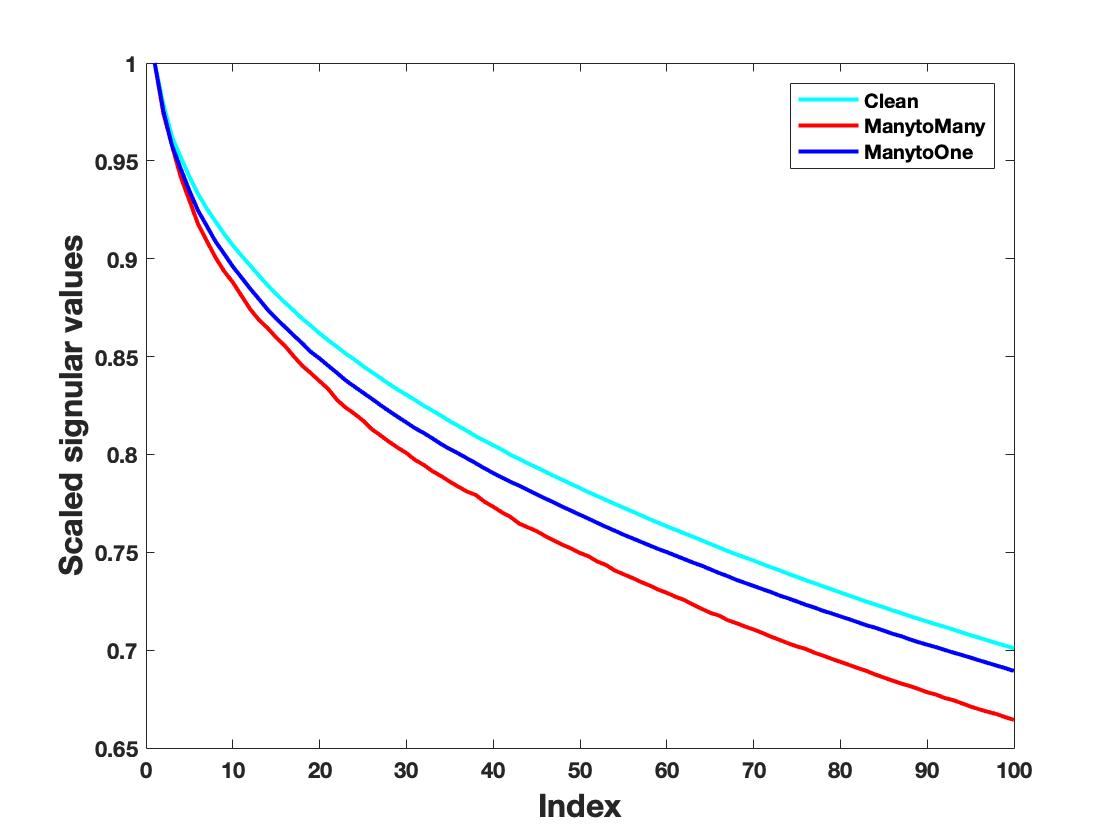}
    \caption{The first 100 singular values of matrix $\mathbf{S}$ scaled by the first singular value $\sigma_{i}/\sigma_{1}$.}
    \label{fig:fnorm_svd}
    \vspace{-2mm}
\end{figure}

\begin{figure}[t]
    \centering
    \includegraphics[width=.78\linewidth]{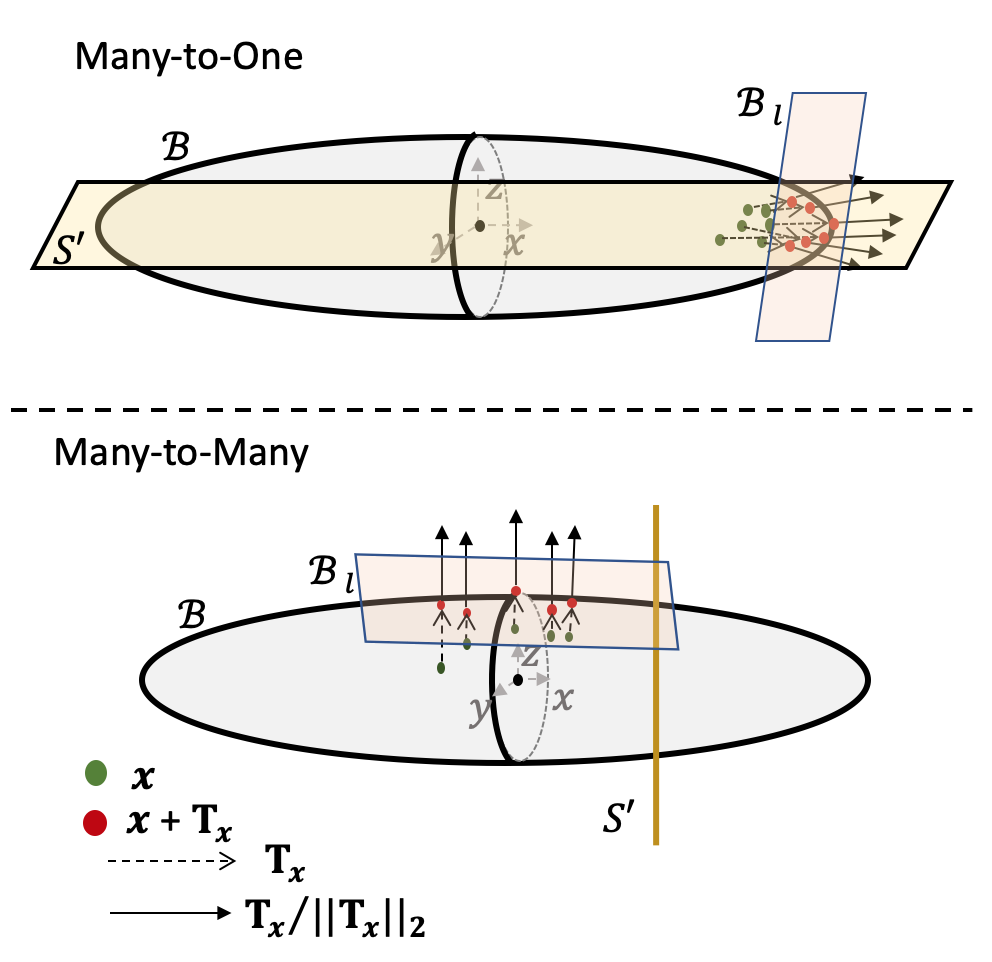}
    \caption {Top: Decision boundary $\mathcal{B}$ of a Trojan model with \textit{M2O} label mapping. The normal vectors to $\mathcal{B}$ (solid arrows) are aligned with the subspace $\mathcal{S}^{'}$ with the dominant direction along x-axis. 
    Bottom:  Decision boundary $\mathcal{B}$ of a Trojan model with \textit{M2M} label mapping. The subspace $\mathcal{S}^{'}$ is along the z-axis with normal vectors parallel to it. In both cases, the non-linear $\mathcal{B}$ can be replaced with linear $\mathcal{B}_{l}$ with dominant direction in $\mathcal{S}^{'}$ as its normal.}% \MS{x, y, z labels are very faint, make them dark}}% \textcolor{red}{does $\mathcal{S}^{'}$ lies in x-y plane for M2O and y-z plane for M2M?}}
    \label{fig:subspaces}
    \vspace{-4mm}
\end{figure}

We create matrix $\mathbf{S}$ for each of the clean and Trojan models of CIFAR-10 and Fashoin-MNIST  datasets using 600 and 300 samples per class from the validation set, respectively.
\textcolor{black}{
Figure \ref{fig:fnorm_svd} shows the distribution of the first 100 singular values based on the label mapping. For ease of comparison, we scale all singular values with the first one. Now, each singular value represents the importance of that coordinate axis compared to the first coordinate axis. The analysis of the distribution of singular values reveals the following findings: \textbf{(I):} The space $\mathcal{S}$ has a significantly lower dimension than $d$ i.e.~$dim(\mathcal{S}) \ll d$ \textbf{II):} The first few singular values have a similar energy pattern in all type of models. However, in the Trojan models regardless of the mapping type, the contribution of the remaining singular values in the total energy of the space $\mathcal{S}$ decreases \emph{more rapidly} compared to clean models. Note that in Figure \ref{fig:fnorm_svd} the red (Many-to-Many mapping) and blue (Many-to-One mapping) curves are consistently below the curve of clean models. This suggests that for the Trojan models, the normal vectors are more aligned with each other and also with the subspace $\mathcal{S}^{'}$ created by the basis correspond to the dominant singular values of $S$. In other words, \textit{Trojan insertion creates a dominant direction in $\mathcal{S}^{'}$}. %\textcolor{red}{(dominant direction for what?. samples x? or normal vector? did we define the term dominant direction? )} 
Figure \ref{fig:subspaces} shows the schematic representation of normal vectors to the decision boundary $\mathcal{B}\in \mathbb{R}^{3}$ along with the corresponding subspace $\mathcal{S}^{'}$ for Many-to-One (M2O) and Many-to-Many (M2M) mappings.
}\textcolor{black}{In \textit{M2O} mapping, the subspace $\mathcal{S}^{'}$ is the x-y plane with x-axis as the dominant direction. %\textcolor{red}{Which seems to be directing towards the nonlinear part of the decision boundary.??} 
For \textit{M2M} mapping, the subspace $\mathcal{S}^{'}$ is along the z-axis and the normals are parallel to it.  %\textcolor{red}{Which seems to be directing towards the linear part of the decision boundary.??}.
%\textcolor{red}{Furthermore, due to the alignment/uniform direction of the normal vectors, we can approximate a linear decision boudary $\mathcal{B}_{l}$!!} 
The $\mathcal{B}_{l}$ is a linear decision boundary that can replace the $\mathcal{B}$ with the dominant direction in $\mathcal{S}^{'}$ as its normal vector.} 
%For the ease of the comparison, we normalzied the singular values based on the Forbenius norm of the matrix $\mathbf{\Sigma}$. Now, each of the singular values represents the contribution of each coordinate axis to the total energy of space $\mathcal{S}$ For instance, the first singular value for the many-to-many mapping has $6.9\%$ energy. 
%} 
%AJMAL: The next two paragraphs have contradictory results. Did we remove the plot for these results from Fig. 5?
\textbf{III):} Trojan insertion can affect the non-linearity of decision boundary differently based on the type of attack one uses. %type used in CIFAR10 models.
For \textit{M2O} mapping, Trojaning slightly increases the non-linearity of the  decision boundary around the manifold of clean data compared to clean models. \textcolor{black}{The first 100 singular values of \textit{M2O} mapping covers $2\%$ less energy compared to that of clean models.} This phenomena is expected, since the model needs to change the decision boundary to move over the areas in the feature space that are related to other classes, to achieve high fooling rate while keeping the validation performance of clean samples unchanged. 
%\textcolor{red}{This also increases the area in the feature space that is dedicated to the target class, as seen in Figure \ref{fig:3D_m2o}. In this Figure, the output logits for 40 samples per class of a network are visualized in 3D space. The samples of Traget class 7 (hollow blue dots) are spread over larger area even in the 3D space.} 
However, \textit{M2M} mapping slightly decreases the non-linearity of the decision boundary %\textcolor{red}{for the models of CIFAR-10}
\textcolor{black}{with the first 100 singular values covering $3\%$ more energy compared to clean models} . We believe that, in this type of mapping, since each True label only maps to one Target label and the poisoning ratio is small, $15\%-20\%$, the triggered samples act like a regularizer during the training process and decrease the non-linearity of decision boundary, while it increases the margin as shown in \textcolor{black}{Table \ref{tbl:avg_margin}}.
%\MS{So what is the bottom line? Table 1 did not do what we were expecting for M2M and Mixed, then you propose  a method with singular values etc; Is this method able to distinguish between clean and M2M and Mixed? Can you put a Table for this similar to Table 1? This will help the reviewers. I assume you are saying just use Figure 5.}
%\textcolor{red}{The 3D visualization of output logits for 40 samples in Figure \ref{fig:3D_m2m} also shows distinctive clusters for each class compared to a clean model in Figure \ref{fig:clean}.}
\iffalse
\begin{table*}
  \caption{Performance of the proposed Trojan detector on four datasets.} 
  \vspace{-3mm}
  \label{tbl:detector_acc}
  \centering
  \begin{tabular}{lllllccc}
    \toprule
    Dataset &$\xi$&$\rho$&$\amthbf{J}$    & $\delta$    & Precision & Recall & Accuracy\% \\
    \midrule
    NIST & 10 &0.5 &5 &50  &0.924$\pm$0.02  &0.753$\pm$0.02 & 83.40$\pm$0.80    \\
    CIFAR10 &10&0.5&10&  50 & 1.000$\pm$0.00  &0.976$\pm$0.01  & 98.73$\pm$0.58     \\
    MNIST  & 5 & 0.5&10& 50 &  0.818$\pm$0.01   & 0.936$\pm$0.01     & 86.36$\pm$1.11   \\
    FashionMNIST  & 5&0.5&10&50 &  1.000$\pm$0.00 & 0.715$\pm$0.04   &    85.29$\pm$2.23   \\
    \bottomrule
  \end{tabular}
  \label{tbl_5fold_perf}
  \vspace{-5mm}
\end{table*}
\fi
\begin{algorithm}[H]
\caption{Trojan Detector }
% \SetAlgoLined
\begin{algorithmic}[1]
\STATE \textbf{Input:} Validation set $\mathcal{D}_{v}$, classifier $M$, magnitude of the perturbation $\xi$, threshold of error rate for perturbed input batch $\rho$, maximum iteration $\mathbf{J}$, performance threshold $\delta$
 
\STATE \textbf{Output:} Detector decision (Clean / Trojan)%Detector perturbation $\mathbf{r}_{\mathbf{X}}$ 
\STATE \textbf{Step 1:}

\STATE  Select image batch $\mathbf{X}$ randomly from $\mathcal{D}_{v}$

\STATE Initialize  $i\leftarrow 0$, $j\leftarrow 0$, $\mathbf{r}_{\mathbf{X}}\leftarrow 0$
 
\WHILE {$j\leq \mathbf{J}$  and $Err(M(\mathbf{X}+\mathbf{r}_{\mathbf{X}}))\leq \rho$}
 
    \FOR{each image $\mathbf{x}_{i} \in \mathbf{X}$}
        \STATE  compute $\mathbf{t}_{\mathbf{x}_{i}+\mathbf{r}_{\mathbf{X}}}$ using Eq. \eqref{eqe:mtnldb} \footnotesize{$ \triangleleft$ Perturbation that projects ${\mathbf{x}_{i}}+\mathbf{r}_{\mathbf{X}} $ onto the nearest point on  $\mathcal{B}_{linearized}$ Eq. \eqref{c_db}} %liniearized at the point ${\mathbf{x}_{i}}+\mathbf{r}_{\mathbf{X}}$}
        %= Perturbation that projects ${\mathbf{x}_{i}}+\mathbf{r}_{\mathbf{X}} $ onto the nearest $\mathcal{B}_{l}$, liniearized at the point ${\mathbf{x}_{i}}+\mathbf{r}_{\mathbf{X}}$ as computed in Equ. \eqref{eqe:mtnldb}
        
    \ENDFOR
 
    \STATE $\mathbf{r}_{\mathbf{X}} \leftarrow \mathbf{r}_{\mathbf{X}}+ \sum_{i} \frac{\mathbf{t}_{{\mathbf{x}_{i}}+\mathbf{r}_{\mathbf{X}}}}{\|\mathbf{t}_{{\mathbf{x}_{i}}+\mathbf{r}_{\mathbf{X}}}\|_2}$ $~~~~\triangleleft$  normal vector to $\mathcal{B}_{l}$
 
    \STATE $\mathbf{r}_{\mathbf{X}} \leftarrow \xi\frac{\mathbf{r}_{\mathbf{X}}}{\|\mathbf{r}_{\mathbf{X}}\|_{2}}$  $~~~~~~~~~~~~~~~~~~~\triangleleft$ scale the normal vector to magnitude $\xi$

    \STATE $j \leftarrow j+1$
\ENDWHILE
%\end{algorithmic}

\STATE \textbf{Step 2:}
\STATE Create perturbed validation set: $\mathcal{D}_{v}^{'} = \{\mathcal{D}_{v}\backslash \mathbf{X}\}+\mathbf{r}_{\mathbf{X}}$
\IF{$Err(M(\mathcal{D}_{v}^{'}))\geq \delta$}
    %\STATE
    \RETURN{ Trojan} 
\ELSE
    \RETURN{ Clean} 
\ENDIF
%\STATE \textbf{Return} $\mathbf{t}_{\mathbf{x}}$
\end{algorithmic}
\label{Alg3}
\end{algorithm}
%\vspace{-4mm}
%\MS{Is analysis in Figure 4  employed in Algorithm 1?}
\section{Trojan Detector}
\vspace{-2mm}
The detector is inspired by our findings in Section \ref{ch:T_Analysis} that Trojaning can \textbf{(i)} create a dominant direction in the perturbation space  %non-linearity of decision boundary 
around the manifold of clean data; \textbf{(ii)} decrease the average margin compared to clean models.
The first finding implies that the non-linear decision boundary, $\mathcal{B}$, can be better represented by a linearized one, $\mathcal{B}_{l}$, around $Q_{data}$.  
 Since the perturbation directions  $\mathbf{T}_{\mathbf{x}_i}$ that project samples $\mathbf{x}_{i}$ to the closest point on the non-linear decision boundary are more aligned, the normal direction to $\mathcal{B}_{l}$ can be found by considering the directions of fewer samples. %"**rewrite it based on subspace dimensionality and scale***"
The second finding suggests that, if we perturb samples along the normal direction of $\mathcal{B}_{l}$ with a certain magnitude, it causes a higher misclassification rate for Trojan models compared to clean models. 

Our Trojan detector consists of two components. The first one is responsible for finding the normal vector to the best representative linearized decision boundary around a small batch of samples $\mathbf{X}\in Q_{data}$, that is scaled to a given magnitude, $\xi$. The output of first step is the detector perturbation vector $\mathbf{r_{X}}$ that maps $\mathbf{X}$ to the linearized decision boundary of $M$.
In the second step, all the samples in the held-out validation set \(\mathcal{D}_{v}\backslash \mathbf{X}\) are perturbed with the detector perturbation $\mathbf{r_{X}}$ as \(\mathcal{D}^{'}_{v}=\{(\mathbf{x}_{i}+\mathbf{r_{X}},\mathbf{y}_{i})|(\mathbf{x}_{i},\mathbf{y}_{i})\in \mathcal{D}_{v}\backslash \mathbf{X} \}\).% follows:
%\begin{equation}
% \mathcal{D}^{'}_{v}=\{(\mathbf{x}_{i}+\mathbf{%r_{X}},\mathbf{y}_{i})|(\mathbf{x}_{i},\mathbf{%y}_{i})\in \mathcal{D}_{v}  \} .
%\end{equation}
The detector considers the \emph{Error rate} of the model $M$ on samples of $\mathcal{D}^{'}_{v}$, denoted as $Err(M(\mathcal{D}^{'}_{v}))$, to differentiate between clean and Trojan models.  %Hence,  if the the detector can be fo
The detector function $Detector(M)$ labels the model $M$ as \emph{Trojan} if $Err(M(\mathcal{D}^{'}_{v})) \geq \delta$, and label it as \emph{clean} otherwise. Here, $\delta$ denotes the performance threshold of the detector and decides the sensitivity of the detector. The proposed Trojan detector is presented in Algorithm \ref{Alg3}. 
\textcolor{black}{Note that the detector perturbation procedure in Algorithm \ref{Alg3} is inspired by Universal Adversarial Perturbation (UAP) \cite{moosavi2017universal} in the sense that both aim to compute a direction in the perturbation space based on a batch of data, $\mathbf{X}$, that causes the misclassification for all the samples. However, our method is inherently different in how they compute the direction. UAP finds the direction sequentially by aggregating the minimal perturbations that sends the current sample $\mathbf{x}_{i}$ that has been perturbed by UAP perturbation $\mathbf{v}$ to the decision boundary of the classifier. While Algorithm \ref{Alg3} tries to find the normal to the linear decision boundary $\mathcal{B}_{l}$ by emphasizing on the alignment of normal vectors to the classifier decision boundary $\mathcal{B}$ in Trojan models. Since this feature is more prominent in Trojan models, the detector perturbation becomes a stronger attack to Trojan models and leads to larger drop in the accuracy compared to clean models. 
\iffalse
eagreement On the other hand, Algorithm \ref{Alg3} 
emphasizes on the alignment of normals to the classifier's decision boundary and computes the normal to the li
 At each step if the current UAP perturbation $\mathbf{v}$ does not cause misclassification of current sample $\mathbf{x}_{i}$, the algorithm computes the minimal perturbation $\Delta v_{i}$ that sends the current perturbed sample $\mathbf{x}_{i}+\mathbf{v}$ to the decision boundary of the classifier. 
 \fi}
% can be formalized as  
\iffalse
\begin{equation}
\mathit{Detector}(M)=\Bigg\{
\begin{matrix}
1 & Err(M(\mathcal{D}^{'}_{v})) \geq \delta,\\
0 & Err(M(\mathcal{D}^{'}_{v})) < \delta,
\end{matrix}
\end{equation}
\fi

%AJMAL: Transpose this table so it matches Table 1 OR transpose Table 1.
%\textbf{Experiments}:
\iffalse
\begin{table}

  %\vspace{mm}
   \footnotesize
  \centering
  \begin{tabular}{cccc}
    \toprule
    \footnotesize{Dataset}  &\footnotesize{MNIST} &\footnotesize{Fashion-MNIST} &\footnotesize{CIFAR10}\\
    \midrule
    \footnotesize{Celan} &\footnotesize{ 80.2$\pm$2.8 } & \footnotesize{100$\pm$ 0 }& \footnotesize{99.5$\pm$0.5}    \\
    \footnotesize{M2O} & \footnotesize{91.8$\pm$5.1 } & \footnotesize{81.6$\pm$7.8} & \footnotesize{96.1$\pm$2.2}    \\
    \footnotesize{M2M} & \footnotesize{92.7$\pm$2.8}  &\footnotesize{74.7$\pm$6.9}  & \footnotesize{99.4$\pm$1.08}     \\
    \footnotesize{Mixed}  & \footnotesize{ 96.4$\pm$4.7}  & \footnotesize{71.0$\pm$8.3 }    & \footnotesize{97.8$\pm$3.0}   \\
    \bottomrule
  \end{tabular}
    \caption{Accuracy $\%$ of the proposed Trojan detectors on Odysseus for different types of attack.}
    \label{tbl:HectorvsOdy}
  \vspace{-5mm}
\end{table}
\fi

\begin{table}

  %\vspace{mm}
   \footnotesize
  \centering
  \begin{tabular}{ccccc}
    \toprule
    \footnotesize{Dataset}  & \footnotesize{Clean} &\footnotesize{M2O} & \footnotesize{M2M} & \footnotesize{Mixed}\\
    \midrule
    \footnotesize{MNIST} &\footnotesize{ 80.2$\pm$2.8 } & \footnotesize{91.8$\pm$5.1 }& \footnotesize{92.7$\pm$2.8}& \footnotesize{ 96.4$\pm$4.7}    \\
    \footnotesize{FashionMNIST} & \footnotesize{100$\pm$ 0 } & \footnotesize{81.6$\pm$7.8} & \footnotesize{74.7$\pm$6.9} &\footnotesize{71.0$\pm$8.3 }    \\
    \footnotesize{CIFAR10} & \footnotesize{99.5$\pm$0.5}   &\footnotesize{96.1$\pm$2.2}  & \footnotesize{99.4$\pm$1.08}  & \footnotesize{97.8$\pm$3.0}      \\
    %\footnotesize{Mixed}  & \footnotesize{ 96.4$\pm$4.7}  & \footnotesize{71.0$\pm$8.3 }    & \footnotesize{97.8$\pm$3.0}   \\
    \bottomrule
  \end{tabular}
  \vspace{0mm}
\caption{\footnotesize{Accuracy of the proposed Trojan detectors on Odysseus for different true label to target label mappings.} }
\label{tbl:HectorvsOdy}
  %\label{tbl_5fold_perf}
  \vspace{-5mm}
\end{table}
\vspace{-2mm}
\section{Experiments}
\vspace{-2mm}
\textcolor{black}{In this section, we evaluate the quality of the Odysseus dataset followed by the performance and generalizability of the proposed Trojan detector.}

\textcolor{black}{
In the first set of experiments we evaluate the performance of the proposed Trojan detector on our Odysseus dataset. The 5-fold cross validation accuracy of the detector for clean and different label mapping is reported in Table \ref{tbl:HectorvsOdy}. For all parts of Odysseus, we set the error rate threshold $\rho=0.5$ and the maximum iteration $\mathbf{J}=10$. The magnitude of perturbation $\xi$ is set to 5 for gray-scale images of MNIST and Fashion-MNIST and 10 for CIFAR10. Finally, $\mathbf{r_{X}}$ is computed based on 40 samples per class with performance threshold of $\delta=0.5$. As it can be seen, the proposed Trojan detector sets a high baseline on Odysseus even with almost fixed set of hyperparameters. For the analysis of the effect of each parameter on the performance please refer to supplementary material.} 
%-------------------------------------
\begin{table}
 %with $\delta=50\%$.} 
  %\vspace{mm}
   \footnotesize
  %\label{tbl:detector_acc}
  \centering
  \begin{tabular}{lllllccc}
    \toprule
    \footnotesize{Dataset}  & \footnotesize{Precision} & \footnotesize{Recall} & \footnotesize{Accuracy(\%)} \\
    \midrule
    \footnotesize{NIST R-0 \cite{TrojAI0}} & \footnotesize{0.851$\pm$0.05}  &\footnotesize{0.928$\pm$0.02} &\footnotesize{85.00$\pm$3.78}    \\
    
    \footnotesize{NIST R-1 \cite{TrojAI1}} & \footnotesize{0.924$\pm$0.02 } &\footnotesize{0.753$\pm$0.02} & \footnotesize{83.40$\pm$0.80}    \\
    
    \footnotesize{NIST R-2 \cite{TrojAI2}} & \footnotesize{0.79$\pm$7.73 } &\footnotesize{0.730$\pm$0.04} & \footnotesize{72.96$\pm$4.37}    \\
    
    \footnotesize{CIFAR10}  & \footnotesize{1.000$\pm$0.00}  &\footnotesize{0.976$\pm$0.01 } & \footnotesize{98.73$\pm$0.58}     \\%Odysseus-CIFAR10
    \footnotesize{MNIST}  &  \footnotesize{0.818$\pm$0.01  } &\footnotesize{ 0.936$\pm$0.01   }  & \footnotesize{86.36$\pm$1.11}   \\
    \footnotesize{FashionMNIST}  & \footnotesize{ 1.000$\pm$0.00 }& \footnotesize{0.715$\pm$0.04  } &    \footnotesize{85.29$\pm$2.23}   \\
    \footnotesize{ULP-TinyImageNet}&\footnotesize{0.790$\pm$ 0.09} & \footnotesize{0.690$\pm$0.02}& \footnotesize{75.61$\pm$1.38} \\
    \bottomrule
  \end{tabular}
  \vspace{-2mm}
    \caption{\footnotesize{Performance of the proposed Trojan detector. }}
  \label{tbl:tbl_7_dataset}
  \vspace{-3mm}
\end{table}
%------------------------------------
\begin{table}[t]
   \footnotesize
  \centering
  \begin{tabular}{cccc}
    \toprule
    \footnotesize{Method}  &\footnotesize{Precision} & \footnotesize{Recall} & \footnotesize{Accuracy(\%)} \\
    \midrule
    \footnotesize{ULP \cite{kolouri2020universal}} &\footnotesize{ 0.780$\pm$0.33 } & \footnotesize{0.518$\pm$0.36 }& \footnotesize{68.63 $\pm$1.49}    \\
    \footnotesize{STRIP \cite{gao2019strip}} & \footnotesize{0.958$\pm$0.02 } & \footnotesize{0.360$\pm$0.01} & \footnotesize{67.32$\pm$1.31}    \\
    \footnotesize{MNTD \cite{xu2019detecting}}  & \footnotesize{1.000$\pm$0.00}  &\footnotesize{ 0.850$\pm$0.01 }    & \footnotesize{92.50$\pm$0.16}   \\
    \footnotesize{NC \cite{wang2019neural}}  & \footnotesize{ 0.854$\pm$0.02}  &\footnotesize{0.408$\pm$0.01 }& \footnotesize{66.83$\pm$1.76 }  \\
    \textbf{Ours} & \footnotesize{\textbf{1.000$\pm$0.00}}  &\footnotesize{\textbf{0.976$\pm$0.01}}  & \footnotesize{\textbf{98.73$\pm$0.58}}     \\
    \bottomrule
  \end{tabular}
  \vspace{1mm}
    \caption{\footnotesize{Performance of SOTA Trojan detectors on Odysseus-CIFAR10. %The NC performance is reported on M2O mapping and clean models.
    }} 
  \label{tbl:CIFAR10comp}
  \vspace{-2mm}
\end{table}
% \vspace{-3mm}
\textcolor{black}{We also evaluate the effectiveness of the proposed Trojan detector on the two other public datasets namely NIST \cite{TrojAI0}-\cite{TrojAI3} and ULP \cite{kolouri2020universal}. The results are presented in Table \ref{tbl:tbl_7_dataset}. Hyper parameters setting are detailed in supplementary material}.
%\textcolor{green}{***Mahtab send it to supmat: For the NIST round-0 and round-1, we used 40 samples per class to compute the perturbation with maximum number of iteration $\mathbf{J}=10$. The rest of the hyper parameters is the same as Odysseus-CIFAR10. For the NIST round-2 dataset, since the dataset provides limited validation set for each model, we used 5 samples per class to compute the perturbation and set the performance threshold $\delta=15$.}  

 \textcolor{black}{ To benchmark the complexity of our new dataset,} we compare the performance of the sate of the art (SOTA) Trojan detectors on the Odysseus-CIFAR10 in Table \ref{tbl:CIFAR10comp}. The Universal Litmus Pattern (ULP) \cite{kolouri2020universal} and Meta-Neural Trojan Detection (MNTD) \cite{xu2019detecting} are training-based detection methods that train a classifier based on the features extracted from clean and Trojan models. MNTD is a blackbox method that requires many shadow benign and Trojan models to learn the decision boundary of the target model. For a fair comparison with other methods, we use it as a whitebox detector. We use $80\%$ of data for training and evaluate on the rest.  %These methods can benefit from a large dataset like Odysseus-CIFAR10. 
 Even after considering 10 litmus patterns, we believe that the poor performance of ULP is due to its weakness in finding ULP patterns for cross architecture models. MNTD performs significantly better than ULP as a whitebox detector. It's $92.50\%$ accuracy is the second best to our method. Applying MNTD as its original blackbox detection mode drops its performance to $64.16\%$.  %In case of MNTD, to have fair comparison with other methods, we treat with it as a white box detection method instead of % training of meta-classifier requires lots of shadow benign and Trojan models. However, instead of their test models (namely, target models), if we use our models the performance drops significantly, accuracy around $60\%$. We also performed another version of this experiment. In this case, we use $80\%$ of our models for training instead of their shadow models and the rest of the models for validation and testing. We show that performance in Table \ref{tbl_5fold_perf} which is slightly worse than our method.        ***** write the analysis based on the updated results of UMTD****% In the other word, the ULP pattern that is suitable for a specific architecture; Resnet for instance; is not proper for other architectures.
 Strong Intentional Perturbation (STRIP)\cite{gao2019strip} is an online defensive method and assumes that Trojan models are input agnostic in the presence of a trigger.%Ajmal: what does this last sentence mean?
 %It assumes that during testing, the detector has access to both clean and Triggered samples.
 The reason for the poor performance of STRIP is that the image agnostic assumption only holds for \emph{fixed trigger position} Trojan models. While in Odysseus, the Trojan models are trained based on random trigger positions.  Neural Cleanse (NC) %operates on the assumption that with small modifications or triggers, one can change the prediction of a Trojan model to a target label. Therefore, through 
 uses optimization to generate a minimal trigger pattern for each label. In Table \ref{tbl:NISTR012}, we compare the performance of these methods along with our proposed detector on other datasets.
\begin{table}
   \footnotesize
  \centering
  \begin{tabular}{ccccc}
    \toprule
    \footnotesize{Method}   & \footnotesize{NIST R-0}& \footnotesize{NIST R-1}&  \footnotesize{NIST R-2} &\footnotesize{TinyImageNet} \\
    \midrule
    \footnotesize{ULP \cite{kolouri2020universal}} & \footnotesize{62.51 }& \footnotesize{56.87} & \footnotesize{54.00} & \footnotesize{\textbf{96.50}$^{*}$}  \\
    \footnotesize{STRIP \cite{gao2019strip}} &  \footnotesize{N/A} & \footnotesize{N/A}& \footnotesize{N/A } & \footnotesize{48.18}    \\
    \footnotesize{MNTD \cite{xu2019detecting}}  &\footnotesize{ 65.14 }    & \footnotesize{57.50} &\footnotesize{49.00}  &\footnotesize{53.40}   \\
    \footnotesize{NC \cite{wang2019neural}}  & \footnotesize{ 65.02 }& \footnotesize{57.71 }& \footnotesize{ 57.07}  &\footnotesize{67.64}  \\
    \textbf{Ours} & \footnotesize{\textbf{85}} &\footnotesize{\textbf{83.40}}   &\footnotesize{\textbf{ 72.96}}  &\footnotesize{76.61} \\
    \bottomrule
  \end{tabular}
  \caption{\footnotesize{Detection Accuracy (\%) of SOTA Trojan detectors on various datasets. STRIP is not applicable to NIST datasets since there is \emph{no triggered samples} available for them. For TinyImageNet, performance of the ULP\textbf{$^{*}$} is reported based on our re-run. }} 
  \label{tbl:NISTR012}
  \vspace{-3mm}
\end{table}

\begin{table}
 %with $\delta=50\%$.} 
  %\vspace{mm}
   \footnotesize
  %\label{tbl:detector_acc}
  \centering
  \begin{tabular}{lllllccc}
    \toprule
    \footnotesize{Method}  & \footnotesize{CIFAR10-Filter} & \footnotesize{CIFAR10-Noise} & \footnotesize{NIST R-3\cite{TrojAI3} } \\
    \midrule
    \footnotesize{ULP \cite{kolouri2020universal}}  &  \footnotesize{62.85$\pm$2.19} &\footnotesize{60.93$\pm$4.41}  & \footnotesize{53.39$\pm$3.54}   \\
    \footnotesize{MNTD \cite{xu2019detecting}}  & \footnotesize{61.42$\pm$1.41}  &\footnotesize{71.87$\pm$3.05} & \footnotesize{46.60$\pm$0.38}     \\%Odysseus-CIFAR10
    \footnotesize{\textbf{Ours} } & \footnotesize{\textbf{96.92$\pm$3.76}}  &\footnotesize{\textbf{84.60$\pm$6.80}} &\footnotesize{\textbf{61.09$\pm$4.01}}    \\
    \bottomrule
  \end{tabular}
  \vspace{1mm}
    \caption{\footnotesize{Accuracy ($\%$) of the proposed Trojan detector on new scenarios. We employ different type of color filters as triggers for \emph{CIFAR10-Filter} models. For \emph{CIFAR10-Noise}, noise has been used as regularizer during training. \emph{NIST-R3}\cite{TrojAI3} models are adverserally trained.}}
  \label{tbl:Noise_Filter_R3}
  \vspace{-3mm}
\end{table}

Finally we compare the performance of Trojan detectors against three complicated unseen scenarios, namely \emph{new triggers}, \emph{regularized} models and \emph{adversarially trained} models. For new triggers, we train 12 Trojan models per mapping with various filters as Trigger for CIFAR10 dataset. For regularized models, We trained in total 160 clean and Trojan models with noise as the regularizer to make them robust against random perturbations. We also test the detectors against adversarially trained models of  NIST-R3\cite{TrojAI3} dataset. Table \ref{tbl:Noise_Filter_R3} reports the performance for each scenarios. The proposed detector shows higher generalizability compare to other methods in all scenarios and performs well on new triggers with only $2\%$ drops in accuracy compared to known triggers. 
The worst performance of our detector is against adversarially trained models of NIST-R3\cite{TrojAI3} dataset. Considering the small-sized validation set of each model, we could only use 5 samples per class to find the dominant perturbation direction which is not enough to recover the correct direction. Furthermore, we believe that the degradation of performance, in both regularized models and adversarially trained models, is related to the  their effect on the shape of decision boundary. 

\vspace{-1mm}
\section{Conclusion}
 \vspace{-2mm}
 We proposed \emph{Odysseus}, the most diverse public Trojan dataset with more than 3000 models. Our analysis on this dataset shows that increasing the Trigger's size adversely affects fooling rate of Trojan models with  \textit{M2M} and \textit{Mix} label mapping. In addition, analysis of the intrinsic properties of Trojan models revealed that (\textit{M2O}) mapping consistently reduces the average margin and Trojan insertion process creates a dominant direction in the perturbation space. Taking these two properties into consideration, we proposed a Trojan detector that works without any information about the attack or training data and sets a high baseline accuracy; for \emph{Odysseus}. While \emph{Odysseus} is a breakthrough, there are still many aspect of Trojan models that needs further investigation. 
 Effect of data augmentation methods and regularizers on the success of Trojan attacks and also intrinsic properties of Trojan models, behaviour of Trojan classifiers with high resolution input and more output classes, mitigation of Trojan attacks are few to name.
 % We leave the study of the effect of adversarial training on the Trojaning performance and shape of decision boundary for future investigation. 
 \section*{Acknowledge}
 This research was partially supported by Australian Research Council Discovery Grant DP190102443.%The views, findings, opinions, and conclusions contained herein are those of the authors and should not be interpreted as necessarily representing the official policies or endorsements, either expressed or implied of Australian Research Council Discovery.

\small
\bibliographystyle{ieee_fullname}
\bibliography{egbib}

\end{document}